\documentclass[final,3p,times]{elsarticle}

\usepackage{amssymb}    
\usepackage{amsmath}    
\usepackage{lipsum}  
\usepackage{graphicx}   
\usepackage{subcaption}
\usepackage{accents}    
\usepackage{array}      
\usepackage{color}      
\usepackage{hyperref}
\hypersetup{
  pdftitle={Esophageal virtual disease landscape using mechanics-informed machine learning},
  pdfauthor={Sourav Halder}
}

\journal{ArXiv}

\begin{document}

\begin{frontmatter}

\title{Esophageal virtual disease landscape using mechanics-informed machine learning}

\author{Sourav Halder\fnref{LB1}}
\author{Jun Yamasaki\fnref{LB2}}
\author{Shashank Acharya\fnref{LB2}}
\author{Wenjun Kou\fnref{LB3}}
\author{Guy Elisha\fnref{LB2}}
\author{Dustin A. Carlson\fnref{LB3}}
\author{Peter J. Kahrilas\fnref{LB3}}
\author{John E. Pandolfino\fnref{LB3}}
\author{Neelesh A. Patankar\corref{cpau}\fnref{LB1,LB2}}
\cortext[cpau]{Corresponding author}
\ead{n-patankar@northwestern.edu}

\address[LB1]{Theoretical and Applied Mechanics Program, Northwestern University, Evanston, IL 60208, USA}
\address[LB2]{Department of Mechanical Engineering, Northwestern University, Evanston, IL 60208, USA}
\address[LB3]{Division of Gastroenterology and Hepatology, Feinberg School of Medicine, Northwestern University, Chicago, IL 60611, USA}

\begin{abstract}

The pathogenesis of esophageal disorders is related to the esophageal wall mechanics. Therefore, to understand the underlying fundamental mechanisms behind various esophageal disorders, it is crucial to map the esophageal wall mechanics-based parameters onto physiological and pathophysiological conditions corresponding to altered bolus transit and supraphysiologic IBP. In this work, we present a hybrid framework that combines fluid mechanics and machine learning to identify the underlying physics of the various esophageal disorders and maps them onto a parameter space which we call the virtual disease landscape (VDL). A one-dimensional inverse model processes the output from an esophageal diagnostic device called endoscopic functional lumen imaging probe (EndoFLIP) to estimate the mechanical “health” of the esophagus by predicting a set of mechanics-based parameters such as esophageal wall stiffness, muscle contraction pattern and active relaxation of esophageal walls. The mechanics-based parameters were then used to train a neural network that consists of a variational autoencoder (VAE) that generates a latent space and a side network that predicts mechanical work metrics for estimating esophagogastric junction motility. The latent vectors along with a set of discrete mechanics-based parameters define the VDL and form clusters corresponding to the various esophageal disorders. The VDL not only distinguishes different disorders but can also be used to predict disease progression in time. Finally, we also demonstrate the clinical applicability of this framework for estimating the effectiveness of a treatment and track patient condition after a treatment.
\end{abstract}


\end{frontmatter}

\section{Introduction} \label{sec:intro}
Medical diagnosis often involves the use of various diagnostic technologies that measure physical quantities such as pressure, velocity of transported fluids, or visualize the geometry and deformation of various tissues.  Although these physical quantities are useful identifiers of various diseases, they are usually not the fundamental physio-markers that define the physical and functional state of an organ like the tissue properties and neural activation, respectively. But clinical decisions are often made based on qualitative trends of these physical quantities instead of the actual physio-markers. This can lead to discrepancies in medical predictions. To mitigate these limitations, it is crucial to consider more quantitative approaches for medical diagnosis by predicting the fundamental physio-markers. Experimental and computational frameworks that combine the output from the diagnostic devices and the physical laws that govern the physical quantities measured by the devices to estimate the physio-markers could potentially lead to better and more accurate clinical decisions. Identifying patterns of similarities and dissimilarities between these physio-markers can provide fundamental insights about the physical nature of different diseases. In this work, we present a novel hybrid approach that uses machine learning and fluid mechanics to process the raw data generated from esophageal diagnostic devices to predict a set of mechanics-based parameters as fundamental physio-markers for esophageal disorders in a patient-specific manner and identifies patterns in these parameters unique to the different categories of these disorders.
\begin{figure}
  \centerline{\includegraphics[scale=0.4]{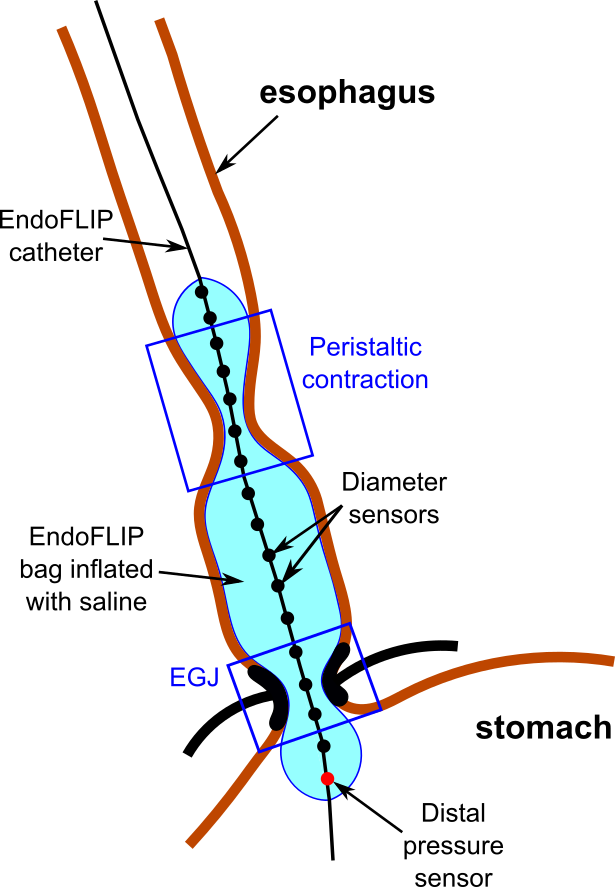}}
  \caption{Schematic diagram of the EndoFLIP. The peristaltic contraction and EGJ tone are indicated inside the blue boxes. Although the EGJ is a complex combination of the lower esophageal sphincter and the diaphragm, it is simply represented here with the LES and diaphragm.}
  \label{fig-endoflip_BCs}
\end{figure}  
One of the most widely used techniques for the diagnosis of esophageal disorders is high-resolution manometry (HRM) \cite{a1,a2,a3,a4,a5}. HRM measures the pressures developed in the esophageal lumen due to the contraction of the esophageal muscles and well as the tone at the esophagogastric junction (EGJ). Clinical diagnosis using HRM is done according to the Chicago Classification (CC) \cite{a6}, which is a hierarchical classification scheme that categorizes the different esophageal disorders. A relatively recent technology for diagnosis of esophageal disorders is the Endoluminal Functional Lumen Imaging Probe (EndoFLIP) \cite{a7,a8}. Fig. \ref{fig-endoflip_BCs} shows the schematic diagram of the EndoFLIP. It consists of a catheter with 16 impedance planimetry sensors to measure the esophageal lumen diameter along its length and a pressure sensor at its distal end. The catheter is mounted with a balloon that can be distended with saline. For measurements, the EndoFLIP is passed trans-orally and placed at the EGJ, and the balloon is distended. This induces repetitive antegrade contractions (RACs), which are unique patterned motor responses to sustained esophageal distension. Carlson et al. \cite{a9} have shown that EndoFLIP can be used to detect esophageal contractility in some Achalasia patients which was not observed in HRM. We used the pressure and diameter data obtained from EndoFLIP to calculate mechanics-based parameters such as esophageal wall properties, muscle contraction strength, effective EGJ tone, and active relaxation of the esophageal muscles. Esophageal biomechanics have been extensively studied using both experimental \cite{a10,a11,a12,a13,a14,a15} and computational \cite{a16,a17,a18,a19,a20,a21,a22,a23,a24} approaches. For our analysis, we used the mathematical framework as described in Halder et al. \cite{a19} to calculate the mechanics-based parameters since it works on clinical diagnosis data from flexible tubular organs and makes fast predictions with limited computational resources.
The mechanics-based parameters quantitatively estimate the mechanical “health” of the esophagus in a patient-specific manner but identifying patterns in these parameters that are unique to various esophageal disorders is a challenging task, especially with the parameters that are functions of both time and location along the esophageal length. It is also crucial to perform this task in a manner that is not sensitive to errors introduced through the EndoFLIP device operation (for instance, irregularities in the placement of the EndoFLIP at the EGJ) as well as to avoid errors and discrepancies introduced due to human intervention (for instance, manually choosing specific values for the mechanics-based parameters that are functions of both time and location along the esophagus. This is analogous to choosing landmarks in pressure topography generated by HRM). This challenging task is tackled using machine learning. Machine learning techniques have been extensively used in medical diagnosis \cite{a25,a26,a27,a28,a29,a30,a31,a32,a33,a34}. They have been used for both medical image analysis and raw patient data analysis. In gastroenterology, however, machine learning has been used mainly for image segmentation and classification tasks \cite{a35,a36,a37,a38}. A recent study \cite{a39} demonstrated the use of a variational autoencoder (VAE) \cite{a40} to identify patterns of various esophageal disorders from raw HRM data. The clusters generated in the latent space of the VAE, therefore, categorizes the raw HRM data into different groups corresponding to the different esophageal disorders. Although the clustering of raw HRM data is beneficial for diagnosis, they do not have a physical meaning. In this work, we present a novel framework, called mechanics-informed variational autoencoder (MI-VAE), which forms clusters in a parameter space corresponding to different esophageal disorders and these clusters have physical meaning since they were generated from mechanics-based parameters. We call this parameter space the Virtual Disease Landscape (VDL). The MI-VAE also has a generative property by which it can predict the future condition of a patient (in terms of mechanics-based parameters) if past time-series data is available. To the best of our knowledge, this is the first demonstration of future prediction of detailed esophageal motility. Additionally, we present how the VDL can aid in providing optimal directions for treatments of various esophageal disorders as well as test the effectiveness of treatments. With these applications, the MI-VAE has immense potential to aid in the diagnosis and treatment of esophageal disorders.

\section{Methods}
The distention of the EndoFLIP bag induces RACs and its sensors measure the variation of diameter along the distal part of the esophagus (including the EGJ) and the pressure at one point at its distal end. Using these outputs from the EndoFLIP, we estimated the mechanical “health” of the esophagus followed by identifying patterns of different esophageal disorders based on the mechanical state of the esophagus. This was done in two steps: 1) using an inverse model to estimate the mechanical “health” of the esophagus by calculating parameters such as esophageal wall properties, contraction strength, active relaxation, work done while opening the EGJ (EGJW) and work required to open the EGJ (EGJROW) \cite{a41}, and 2) using the calculated mechanics-based parameters as input to a VAE which generates the VDL in the form of its latent space. The next two subsections discuss these steps in details.

\subsection{Mechanics model} \label{sec:ib_method_info}
\begin{figure}
  \centerline{\includegraphics[width=\textwidth]{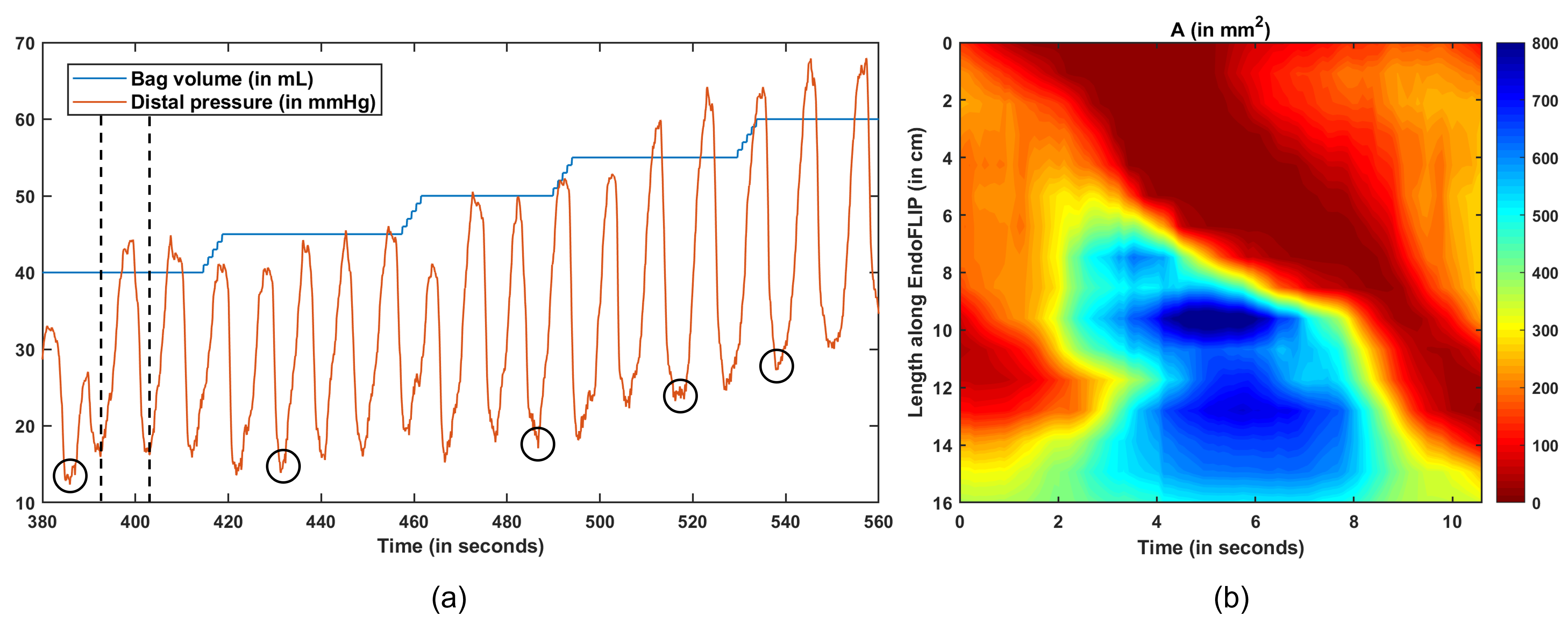}}
  \caption{Output from the EndoFLIP. (a) Bag volume and distal pressure measured by the EndoFLIP catheter. The pressure variations are caused due to secondary peristalsis. The black circles show the local minimum distal pressure. These low-pressure points are assumed to correspond to $\theta\thicksim1$, (b) Variation of cross-sectional area of the EndoFLIP bag in the time interval marked by the dashed lines in (a). The EGJ is seen to distend as the peristaltic contraction moves from the proximal to the distal end of the EndoFLIP.}
  \label{fig-EndoFLIP_output}
\end{figure}
Fig. \ref{fig-EndoFLIP_output} shows a typical output from EndoFLIP. During an EndoFLIP test, the EndoFLIP bag volume is kept constant for while which induces RACs that move from the proximal to the distal end of the EndoFLIP. This leads to a rise in pressure in the distal part of the EndoFLIP as recorded by the distal pressure sensor (red curve in Fig. \ref{fig-EndoFLIP_output}(a)). The 16 impedance sensors record the variation of diameter as a function of time and length along the EndoFLIP. The red band in Fig. \ref{fig-EndoFLIP_output}(b) shows the movement of an antegrade contraction from the proximal to the distal end of the EndoFLIP, and this leads to an increase in diameter at the EGJ shown by the blue region near the distal end. The output from the EndoFLIP consists of diameters measured along its length and pressure measured at only one point. Therefore, there is a lack of information about the three-dimensional geometry of the esophageal lumen and the measured diameters are an approximation in one-dimension i.e., along its length. Additionally, the parameters calculated by the mechanics-based model are required to be done in patient-specific manner. Therefore, it is necessary to use a model that can predict with reasonable accuracy on limited computational resources and time. To that extent, we modeled the EndoFLIP as a one-dimensional flexible tube.
\begin{figure}
  \centerline{\includegraphics[width=\textwidth]{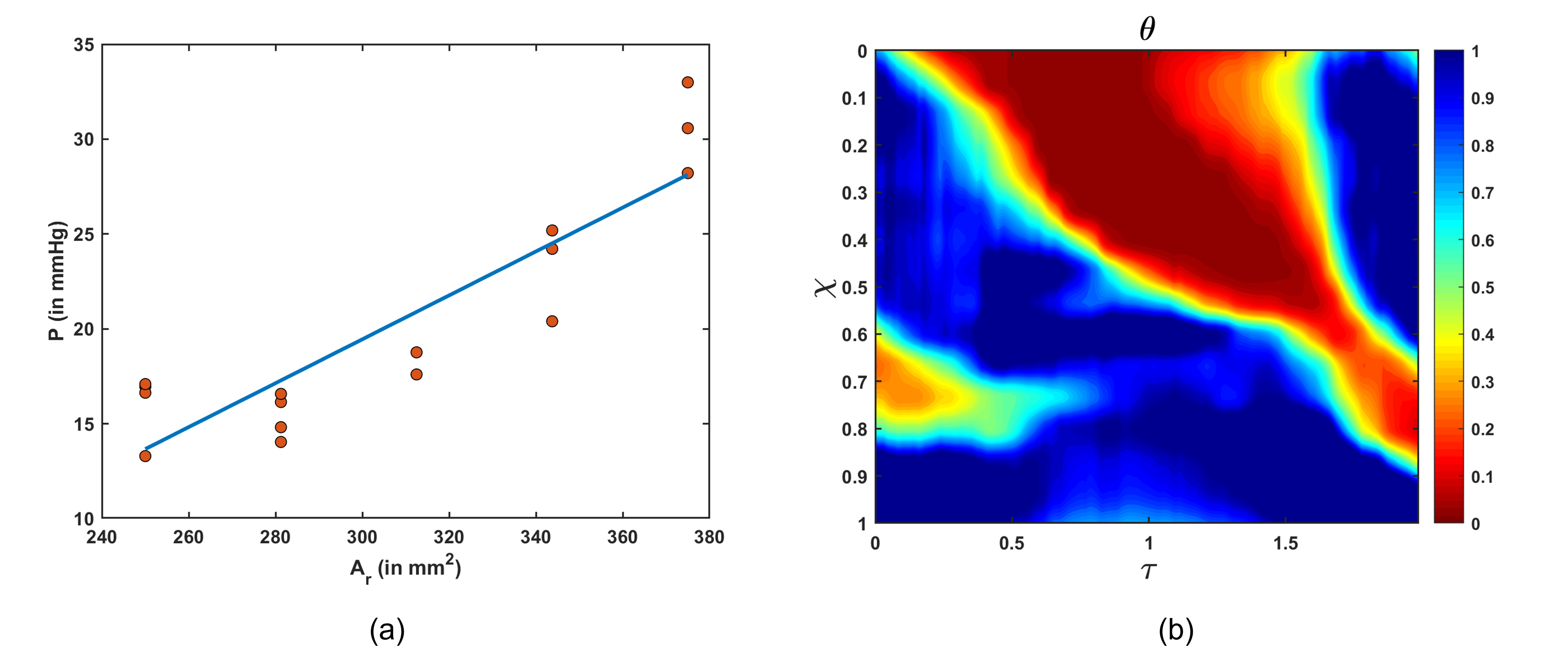}}
  \caption{Mechanics-based parameters. (a) The variation distal pressure with the reference cross-sectional area. The slope and y-intercept of the fitted straight line through linear regression gives $K/A_o$ and $P_o-K$, respectively. (b) Variation of the activation parameter corresponding to the cross-sectional area variation shown in Fig. \ref{fig-EndoFLIP_output}(b).}
  \label{fig-KbyAo_theta}
\end{figure}
\subsubsection{Governing equations}
The one-dimensional mass and momentum equations that describe fluid flow through a flexible tube \cite{a42,a43,a44,a45} are as follows:
\begin{align}
 \frac{\partial A}{\partial t} + \frac{\partial AU}{\partial x} =0, \label{eqn-continuity} \\
 \frac{\partial U}{\partial t} + \frac{\partial}{\partial x}\left(\frac{U^2}{2}\right)+\frac{1}{\rho}\frac{\partial P}{\partial x} + \frac{8\pi\mu U}{\rho A} = 0, \label{eqn-momentum}
\end{align}
where $A$ is the cross-sectional area of the EndoFLIP,  $U$ and $P$ are the velocity and pressure in the fluid inside the EndoFLIP, respectively. $x$ and $t$ represent the distance along the length of the EndoFLIP and time, respectively. $\rho$ and $\mu$ are the density and dynamic viscosity of the fluid, respectively. In Eqns. \ref{eqn-continuity} and \ref{eqn-momentum}, $A$ was known for all $x$ and $t$. The known values of A were used along with Eqns. \ref{eqn-continuity} and \ref{eqn-momentum} to solve for $U$ and $P$.
The fluid pressure inside the esophagus has been found to be linearly proportional to its cross-sectional area \cite{a46,a47}. Hence, it is possible to relate the fluid pressure inside the esophagus with its stiffness as shown below:
\begin{align}
P=P_o+K\left(\frac{A}{\theta A_o}-1\right), \label{eqn-tube_law}
\end{align}
where $K$ is the stiffness of the esophageal walls, $P_o$ is the pressure outside the esophagus (mostly thoracic pressure), $A_o$ is the relaxed cross-sectional area of the esophageal lumen, and $\theta$ is the activation parameter. When the esophagus is in its relaxed state, $\theta=1$. At this state, $A=A_o$, and so, $P=P_o$ i.e., the pressure inside the esophagus is equal to the pressure outside it. When $\theta<1$, a contraction is induced in the esophagus and the pressure rises. $\theta>1$ indicates active relaxation and reduces the pressure inside the esophagus by providing additional space to incorporate a swallowed bolus.

\subsubsection{Boundary conditions}
The EndoFLIP in operation is basically a flexible tube closed at its two ends. In our one-dimensional model of the EndoFLIP, $U=0$ at $x=0$ and $x=L$, where $L$ is the length of the EndoFLIP. Additionally, with the pressure measured by the distal pressure sensor $(P_d)$, $P=P_d (t)$  at  $x=L$.

\subsubsection{Calculation of the primary mechanics-based parameters}

The non-dimensional form of Eqns. \ref{eqn-continuity}-\ref{eqn-tube_law} can be written as follows:
\begin{align}
 \frac{\partial \alpha}{\partial \tau} + \frac{\partial \alpha u}{\partial \chi} =0, \label{eqn-nd_continuity} \\
 \frac{\partial u}{\partial \tau} + \frac{\partial}{\partial \chi}\left(\frac{u^2}{2}\right)+\frac{\partial}{\partial \chi}\left(\frac{\alpha}{\theta}\right) + \gamma\frac{u}{\alpha} = 0, \label{eqn-nd_momentum}
\end{align} 
where  $\chi=x/L$, $\tau=ct/L$, $u=U/c$, $\alpha=A/A_s$ , $p=P/K$, $A_s=(\rho c^2 A_o)/K$, and $\gamma=8\pi\mu L/(\rho^2 c^3 ) (K/A_o )$. Eq. \ref{eqn-tube_law} was used to replace $P$ in Eq. \ref{eqn-momentum} and non-dimensionalized to obtain Eq. \ref{eqn-nd_momentum}. The velocity scale c was taken to be 3 cm /s which is typically the speed of a peristaltic contraction. The scale for non-dimensionalizing cross-sectional area was $A_s$. The viscosity parameter $\gamma$ depends on the ratio of esophageal stiffness $(K)$ and relaxed cross-sectional area $(A_o)$. The three unknown parameters of Eq. \ref{eqn-tube_law}: $K$, $\theta$, and $A_o$ cannot be solved directly due to lack of enough equations. Therefore, we take an indirect approach as described in Acharya et al. \cite{a41}. The ratio $K/A_o$ can be considered as a measure of the stiffness of the esophageal walls. We calculated $K/A_o$ and $P_o-K$ by finding the slope and intercept, respectively, of a fitted line in the plot of $P$ vs. $A$ as shown in Fig. \ref{fig-KbyAo_theta}(a) according to Eq. \ref{eqn-tube_law}. This could be done only by selecting $P$ and $A$ where $\theta=1$. Usually the time instants at which the distal pressure $(P_d)$ are recorded the lowest values for every bag volume, correspond to the time instants when it is reasonable to assume $\theta\thicksim 1$ as shown in Fig. \ref{fig-EndoFLIP_output}(a). In order to select the correct cross-sectional areas at these time instants in an automated manner and also avoid local regions of active relaxation where $\theta>1$, we use a reference cross-sectional area $A_r$ calculated as follows:
\begin{align}
A_r=\frac{V}{L}, \label{eqn-ref_area}
\end{align}
where $V$ is the EndoFLIP bag volume and $L$ is the length of the EndoFLIP. $A_r$ is the cross-sectional area of the EndoFLIP bag when it takes the shape of a perfect cylinder. With $K/A_o$ and $(P_o-K)$ known, and $P$ known at the distal end, we calculated the boundary condition for $\theta$ in Eq. \ref{eqn-nd_momentum} using the relation in Eq. \ref{eqn-tube_law}. 

Eqns. \ref{eqn-nd_continuity} and \ref{eqn-nd_momentum} were solved using the numerical approach described in Halder et al. \cite{a19}. The details of the numerical solution are provided in the Supplementary. In general, the mechanics model described above works on EndoFLIP data for specific time intervals of recordings for cross-sectional area and pressure variations, for instance, the time taken for an antegrade contraction to travel along the EndoFLIP length and merge with the EGJ tone (as shown in Fig. \ref{fig-EndoFLIP_output}(b)). In summary, the primary mechanics-based parameters which are either the input or output in the mechanics model described above are stiffness measure $(K/A_o)$, estimate of the pressure outside the esophagus $(P_o-K)$, maximum distal pressure recorded during a time interval of interest $(P_{\mathrm{max}})$, time taken by an antegrade contraction to pass over the EndoFLIP $(T_{\mathrm{max}})$, EndoFLIP bag volume $(V)$, and the contraction and relaxation pattern described through the activation parameter $(\theta(x,t))$. The activation parameter estimates the contraction strength, the EGJ tone as well as the active relaxation of the esophageal walls.  The active relaxation can be estimated through the maximum value of $\theta$ in the time interval analyzed and we represent it by $\theta_{\mathrm{max}}$, which was considered separately as another primary mechanics-based parameter. We call these parameters primary since they completely define the mechanical state of the esophagus, and other mechanics-based parameters can be calculated directly using a combination of this set of primary parameters.

\subsubsection{Calculation of the secondary mechanics-based parameters: EGJ work metrics}
The EGJ work metrics as described in Acharya et al. \cite{a41} were considered for the secondary parameters. The first EGJ work metric i.e., the work done in opening the EGJ is defined as follows:
\begin{align}
EGJW=\int_{t_1}^{t_2} \int_{x_1}^{x_2} P\frac{\partial A}{\partial t} dxdt, \label{eqn-egjw}
\end{align}
where $t_1$ and $t_2$ are the time instants between which the EGJ opens from its lowest to highest diameter, and $x_1$ and $x_2$ are the proximal and distal bounds of the EGJ along the esophageal length. The second EGJ work metric i.e. the work required to open the EGJ is defined as follows:
\begin{align}
EGJROW=\int_{A_1}^{A_2}\left[\left(\frac{K}{A_o\theta}\right)A+\left(P_o-K\right)\right]\left(x_2-x_1\right)dA, \label{eqn-egjrow}
\end{align}
where $A_1$ and $A_2$ are the lowest and highest reference cross-sectional areas at the EGJ as measured on EndoFLIP. The cross-sectional areas $A_1$ and $A_2$ correspond to the diameters 3 mm and 22 mm, respectively. Since $\theta$ is a function of both $x$ and $t$, EGJROW is also a function of $x$ and $t$. For simplicity, we choose three values of EGJROW corresponding to: 1) $\theta$ calculated at $t_1$ ($EGJROW_1$), 2) median value of $\theta$ calculated between $t_1$ and $t_2$ ($EGJROW_2$), and 3) minimum value of $\theta$ calculate between $t_1$ and $t_2$ ($EGJROW_3$). 

\subsection{Mechanics-informed variational autoencoder}
The mechanics-based parameters calculated by the 1D inverse model give a quantitative estimate of the mechanical “health” of the esophagus through the wall mechanical properties as well as the esophageal motility. Identifying similarities and dissimilarities of these parameters across various esophageal disorders is a crucial step in the development of the VDL. This was done in an unsupervised manner with the help of a variational autoencoder (VAE). Since this neural network works entirely on the mechanics-based parameters, we call it the mechanics informed variational autoencoder (MI-VAE).

\subsubsection{Network architecture}
\begin{figure}
  \centerline{\includegraphics[width=\textwidth]{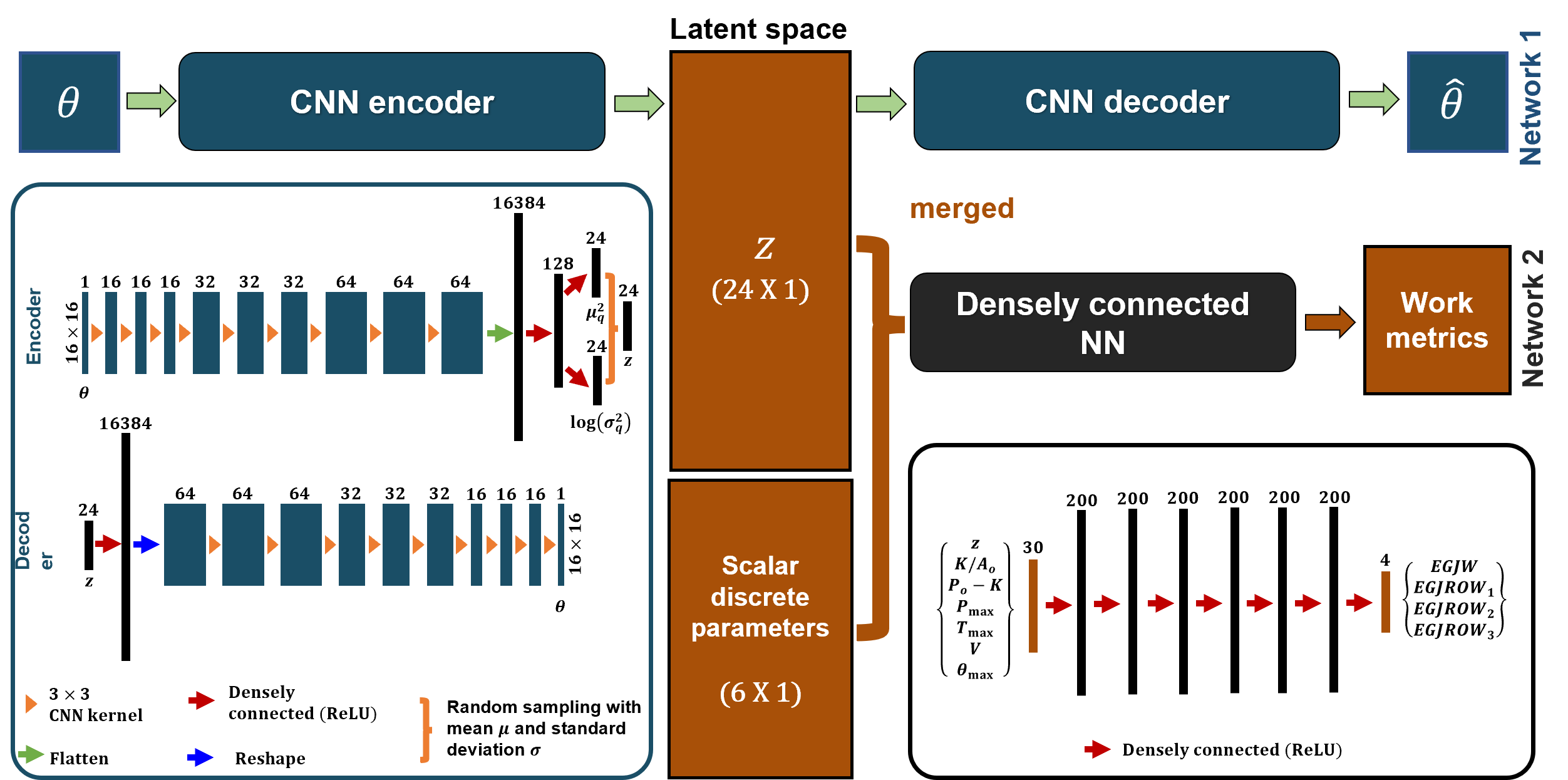}}
  \caption{Network architecture of the mechanics-informed variational autoencoder. In the architecture for network 1, the numbers on the top of the boxes represent the number of channels and the numbers and the output size is represented on the side. For network 2, the numbers on the top represent the number of hidden units.}
 \label{fig-network_architecture}
\end{figure}
The mechanics-based parameters are as follows: $\theta(x,t)$, $K/A_o$, $P_o-K$, $P_{\mathrm{max}}$, $T_{\mathrm{max}}$, $V$, $\theta_{\mathrm{max}}$, $EGJW$, and the 3 measures of $EGJROW$. All these parameters are scalar values except $\theta$, which varies with $x$ and $t$. Thus, simple statistics can be used to identify the patterns of these discrete scalar quantities. But identifying patterns quantitatively with the activation parameter $\theta$, which describes the esophageal motility, requires a different approach. Since the variation of $\theta$ takes the form of a matrix, as shown in Fig. \ref{fig-KbyAo_theta}(b), we used a convolutional neural network based VAE, which we called network 1, to identify the unique patterns of various esophageal disorders through the generated latent space. We used a latent space of 24 dimensions. We used ReLU as the activation function for all the layers of network 1. Additionally, we merged a 6-dimensional vector consisting of a set of discrete parameters to the 24-dimensional vectors generated in the latent space of network 1. These combined 30-dimensional vectors populate a parameter space that forms the VDL. The details of the network architecture are shown in Fig. \ref{fig-network_architecture}.  
The combined 30-dimensional vector, generated with the latent space of network 1 and the vector of discrete parameters, became input to a second neural network, which we called network 2, to predict EGJW and the 3 measures of EGJROW. Network 2 consisted of a densely connected neural network of 3 hidden layers with 75 hidden units each. We used ReLU as the activation function for all the layers of network 2.  Network 1 and 2 together form the MI-VAE that develops the VDL and give physical significance to the vectors in the VDL. 
The work metrics, EGJW and EGJROW, were not included in the 30-dimensional vectors of VDL but kept as separate entities to be predicted by network 2. This was done so that the VDL was developed completely from primary mechanics-based parameters. These primary parameters completely define the mechanical stale and functioning of the esophagus. The work metrics are secondary mechanical parameters that can be derived from the fundamental mechanics-based parameters, and so, they have been kept separate to avoid unnecessary biasing of VDL. The network 2 not only predicts the work metrics, but also forms a framework to test different derivable parameters that can potentially be used as physio-markers for various esophageal disorders.  

\subsubsection{Data}
\begin{table}
\centering
\caption{Cohort details that contributed to training MI-VAE}
\begin{tabular}{|l|c|}
\hline
\textbf{Group}		&		\textbf{Number of subjects}		\\ \hline\hline
Normal	&		237		\\ \hline
Type I achalasia		&		76		\\ \hline
Type II achalasia		&		148		\\ \hline
Type III achalasia		&		47		\\ \hline
EGJ outflow obstruction (EGJOO)		&		27		\\ \hline
Hypercontractility (HC)		&		13		\\ \hline
Distal esophageal spasm (DES)		&		11		\\ \hline
Ineffective esophageal motility (IEM)		&		44		\\ \hline
Absent contractility (AC)		&		17		\\ \hline
Eosinophilic esophagitis (EoE)		&		45		\\ \hline
Gastroesophageal reflux disease (GERD)		&		11		\\ \hline
Scleroderma (SSc)		&		5		\\ \hline
Inconclusive (Inc)		&		123		\\ \hline
\end{tabular}
\label{tab-cohort_details}
\end{table}
\begin{figure}
  \centerline{\includegraphics[width=\textwidth]{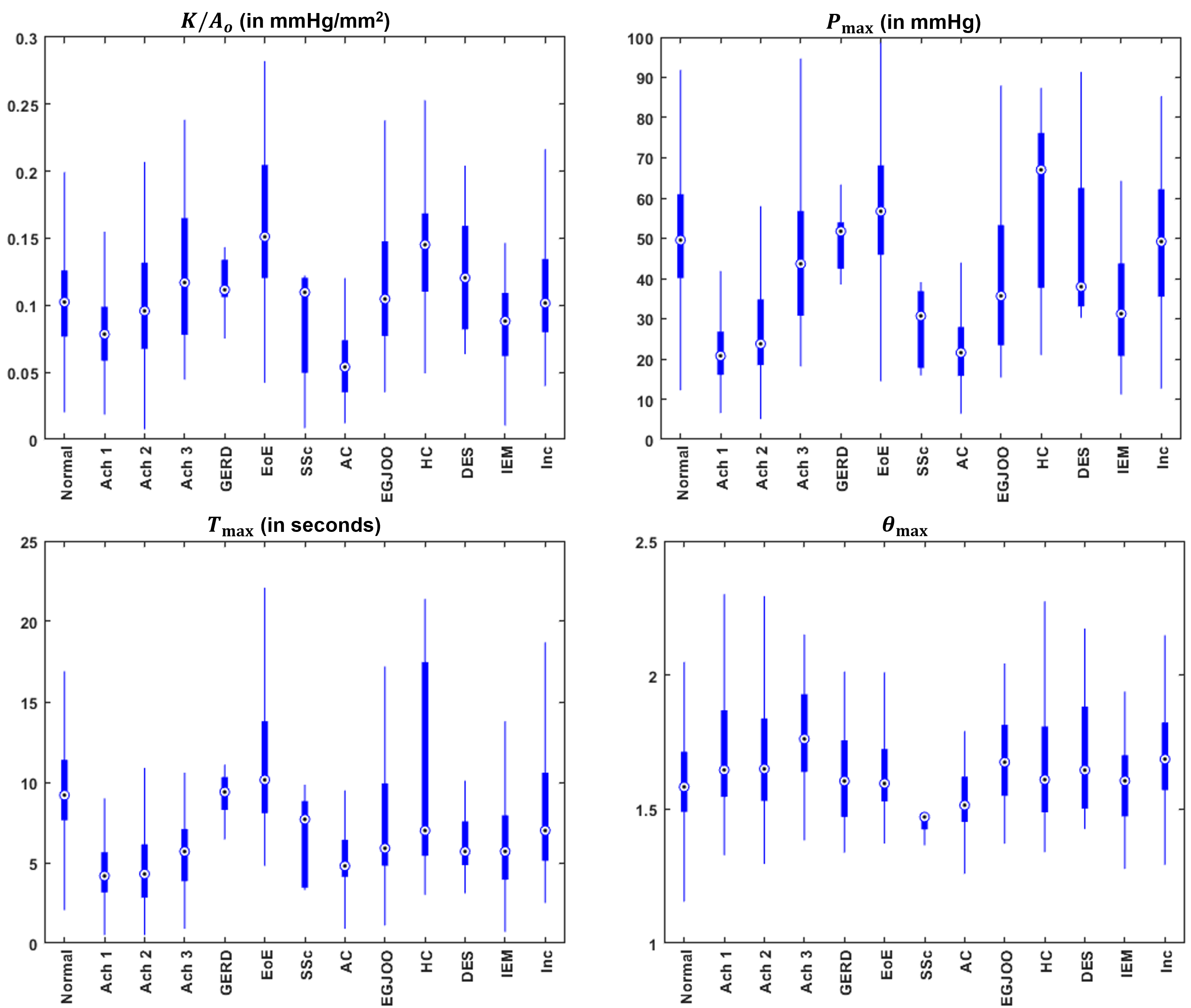}}
  \caption{Box plots showing the distribution of the 4 mechanics-based parameters: $K/A_o$, $P_{\mathrm{max}}$, $T_{\mathrm{max}}$, and $\theta_{\mathrm{max}}$.}
 \label{fig-discrete_parameters}
\end{figure}
\begin{figure}
  \centerline{\includegraphics[width=\textwidth]{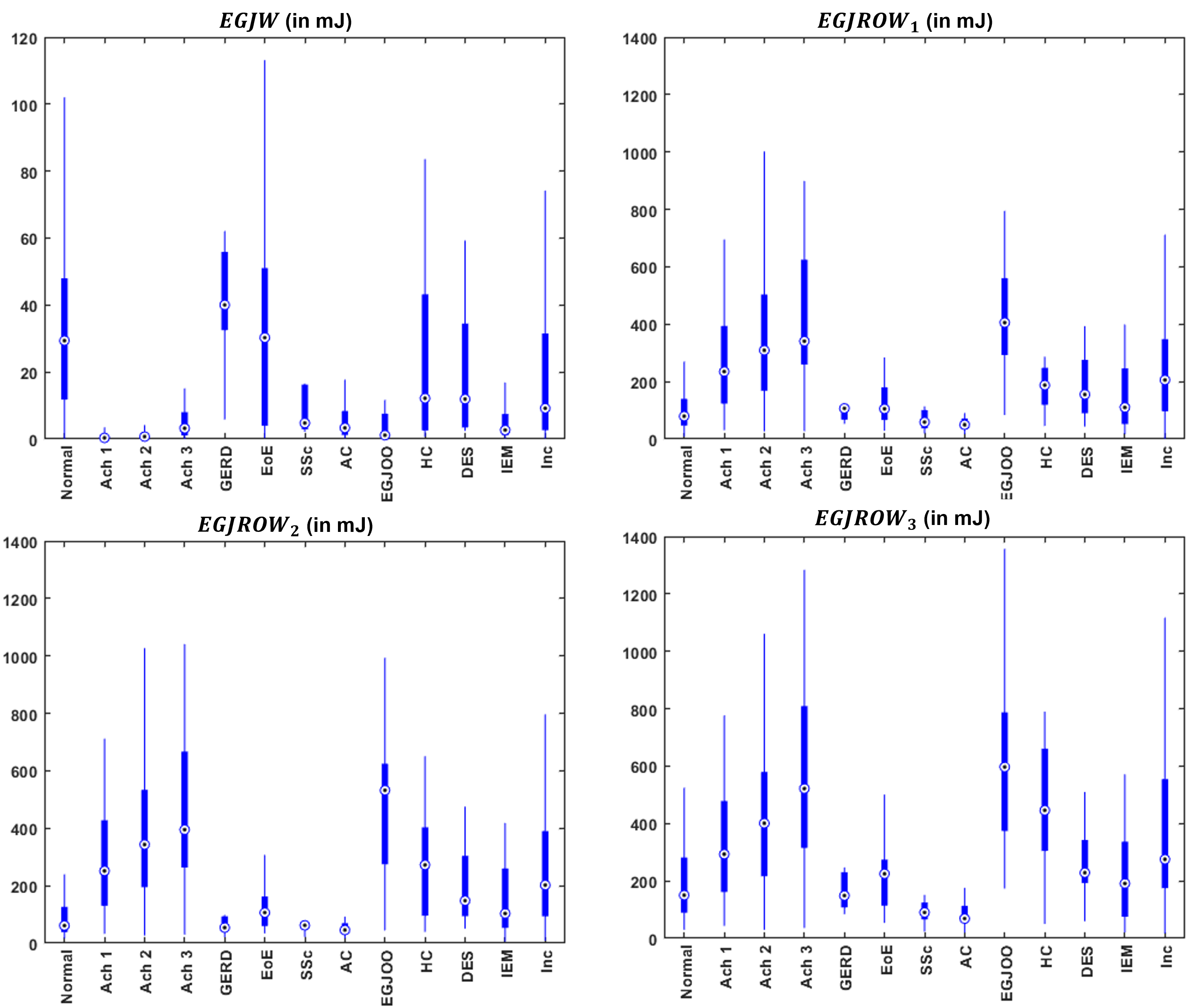}}
  \caption{Box plots showing distribution of the EGJ work metrics.}
 \label{fig-work_metrics}
\end{figure}
The EndoFLIP data used to train the MI-VAE was collected from a cohort of 804 subjects. The details of the cohort are provided in Table \ref{tab-cohort_details}. Both HRM and EndoFLIP tests were performed on these subjects. They were classified into different groups based on their HRM tests using the CC. Of these cohort, 24 Achalasia and 20 EoE patient data were available before and after treatment with topical steroid. Additionally, for one achalasia patient, post-POEM data was available for tracking (1st year, 4th year, and 7th year).  The distributions of the primary and secondary mechanical parameters are shown in Figs. \ref{fig-discrete_parameters} and \ref{fig-work_metrics}, respectively. The work metrics show trends very similar to that reported in Acharya et al. \cite{a41}. 
The patient cohort generated a total dataset of size 7,187. This dataset was augmented to generate a bigger dataset of 222,797.  The parameters $K/A_o$, $P_o-K$, $P_d$, and $T_{\mathrm{max}}$ were augmented by multiplying with a factor $f$ calculated as follows:
\begin{align}
f=1+0.05\mathcal{N}
\end{align}
where $\mathcal{N}$ is a random sampling from a normal distribution with its magnitude less than 2. The raw cross-sectional area variation $A(x,t)$ was augmented using a combination of grid distortional, elastic transformation, and motion blur available in the opensource Python library Albumentations \cite{a48}. The details of this augmentation are provided in Table \ref{tab-augmentation}.
Using the augmented cross-sectional areas and the augmented values for $K/A_o$, $P_d$ and $T_{\mathrm{max}}$, we solved Eqns. \ref{eqn-nd_continuity} and \ref{eqn-nd_momentum} to obtain $\theta(x,t)$. Finally, the corresponding $\theta_{\mathrm{max}}$ and $P_{\mathrm{max}}$ were calculated by taking the maximum of $\theta$ and $P_d$, respectively.
\begin{table}
\centering
\caption{Augmentation details for cross-sectional area}
\begin{tabular}{|l|c|}
\hline
\textbf{Augmentation type}		&		\textbf{Albumentations parameters}		\\ \hline\hline
Grid distortion	&		p=0.9, num\_steps=4, distort\_limit = (-0.2,0.2)		\\ \hline
Elastic transformation		&		p=0.8, alpha=5.0, sigma=100, alpha\_affine = 2.0		\\ \hline
Motion blur		&		p=0.7, blur\_limit = (3,6)		\\ \hline
\end{tabular}
\label{tab-augmentation}
\end{table}

\subsubsection{Training and prediction}
The VAE representing network 1 learns to model an input dataset as a distribution, $p(\theta)$ such that the input variable $\theta$ is generated from a likelihood distribution $p(\theta|z)$, where $z$ is the latent variable. This distribution $p(\theta)$ is parameterized by the weights of the neural network. The decoder yields the likelihood distribution $p(\theta,z)$, i.e., it takes $z$ as input and outputs $\theta$. The encoder should ideally yield the posterior distribution $p(z\vert\theta)$. Unfortunately, $p(z\vert\theta)$ is computationally intractable in general. So, in practice, the encoder yields a conditional distribution $q(z\vert\theta)$ which approximates $p(z\vert\theta)$. The Kullback-Leibler divergence (KLD) provides a measure of the difference between the two distributions $q(z\vert\theta)$ and $p(z\vert\theta)$, and leads to the following relation:
\begin{align}
D_{KL}\left[q\left(z\vert\theta\right)\Vert p\left(z\vert\theta\right)\right]=-\int q\left(z\vert\theta\right)\left[\log p\left(\theta\vert z\right) + \log p(z) -\log q\left(z\vert\theta\right)\right] dz + \log p(\theta). \label{eqn-dkl}
\end{align}
Since KLD is always positive, the right-hand side of the above expression can be written as follows:
\begin{align}
p(\theta) \geq \int q\left(z\vert\theta\right)\left[\log p\left(\theta\vert z\right) + \log p(z) -\log q\left(z\vert\theta\right)\right] dz.
\end{align}
The above equation can be re-written in terms of a new KLD form as follows:
\begin{align}
p(\theta) \geq -D_{KL}\left[q(z\vert\theta)\Vert p(z)\right] + \mathbb{E}_{z\thicksim q(z\vert\theta)} \left[\log p(\theta\vert z)\right], \label{eqn-ptheta}
\end{align}
where $\mathbb{E}_{z\thicksim q(z\vert\theta)} \left[\log p(\theta\vert z)\right]$ is the joint log-likelihood of the input $\theta$ and the latent variable $z$. The right-hand side is called the Evidence Lower bound (ELBO) and is named so since it estimates the lower bound of the likelihood of the data. Thus, maximizing ELBO maximizes the likelihood of the data. The KLD term works like a regulariser and forces the approximate posterior $q(z\vert\theta)$ to be as close to the prior $p(z)$ as possible. This term causes the posterior $q(z\vert\theta)$ to enforce a high probability to $z$ values that can generate the point $\theta$ without collapsing to a single point like an autoencoder. This gives a continuous behavior to the latent space so that meaningful generations are possible from points in the latent space which are not related to any input in the training dataset. The second term of Eq. \ref{eqn-ptheta} is the reconstruction error between in the input and the generated output of the entire network. It is possible to derive a closed form solution for the KLD term if we choose the approximate posterior $q(z\vert\theta)$ to have a Gaussian distribution and choose the prior $p(z)$ to have a standard normal distribution as shown below:
\begin{align}
-D_{KL}\left[q(z\vert\theta)\Vert p(z)\right] = \frac{1}{2}\left[1+\log \sigma_q^2 - \sigma_q^2 - \mu_q^2\right]. \label{eqn-kld}
\end{align}
The encoder, as shown in Fig. \ref{fig-network_architecture}, outputs $\mu_q^2$ and $\log \sigma_q^2$. Although we have an analytical form for the KLD term of Eq. \ref{eqn-ptheta}, the reconstruction error requires to be estimated by sampling. Sampling $z$ from $q(z\vert\theta)$ directly leads to a problem in implementing backpropagation since the network would have a random node at the input of decoder. This problem can be tackled by a reparameterization trick where $z$ is sampled from the mean and log variance parameters of $q(z\vert\theta)$ as estimated by the encoder shown as follows:
\begin{align}
z = \mu_q + \varepsilon\cdot\exp\left(\frac{1}{2}\log\sigma_q^2\right), \label{eqn-latentvar}
\end{align}
where $\varepsilon$ is a random number generated from a standard normal distribution. This step makes it possible to backpropagate in a deterministic manner by considering $\varepsilon$ as an extra input. Since $\varepsilon$ is sampled from a different distribution which is not a function of any variables with respect to which derivatives might be required, stochasticity is introduced in the network without affecting backpropagation. The final form of the loss function used for network 1 is as follows:
\begin{align}
L_{N1} := \frac{1}{2M} \sum_i^M \left[1+\log \sigma_{q,i}^2 - \sigma_{q,i}^2 - \mu_{q,i}^2\right] + \frac{\beta}{N}\sum_j^N \left(\theta_j - \hat{\theta}_j\right)^2, \label{eqn-loss_function}
\end{align}
where $M$ is the dimension of the latent space, $N$ is the product of the two spatial dimensions of the input and generated output. In this case, $M=24$ and $N=256$. The first term is the KLD and the second term is the reconstruction loss. $\beta$ is a scaling parameter used to balance the magnitudes of the reconstruction loss and the KLD for proper training of network 1. We found that $\beta=1000$ resulted in a good balance between the two losses. The loss function shown in Eq. \ref{eqn-loss_function} is a described for each input. While training, we defined the total loss as the mean of $L_{N1}$ calculated over the mini-batch dataset.

We trained network 1 first followed by network 2 with the mean latent space variables and the discrete mechanics-based parameters merged as its input. The input of network 1 i.e., $\theta$, was scaled to lie between 0 and 1, and then subtracted from 1 so that the lower values at the contraction zones would have higher values instead. Thus, network 1 focusses on minimizing the reconstruction error at the contraction zones since they have the most importance in the variation of $\theta$.  Of the 222,797 mechanics-based parameters, 204,000 were used for training and the rest were used for testing.  Network 1 was trained for 250 epochs. We used a learning rate of $1\times10^{-4}$ for the first 100 epochs, $3.3\times10^{-5}$ for the next 100 epochs, and finally $5\times10^{-6}$ for the last 50 epochs. Adam \cite{a49} was used as the optimizer. Network 1 achieved an average mean-squared error between input and generated output as $1.36\times10^{-3}$ and a KLD loss of 1.54.

The input and output of network 2 are shown in Fig. \ref{fig-network_architecture}. The input and output were scaled to lie between 0 and 1. We used a mean-squared error loss function for network 2. Of the total dataset, 2048 were used for validation and the rest for training. To train this network, we used a learning rate of $10^{-3}$ and used RMSprop as optimizer. Network 2 was trained for 1000 epochs, and it achieved a mean-squared error accuracy of $1.88\times10^{-5}$ on the validation dataset. We used Keras \cite{a50} running on top of TensorFlow \cite{a51} to train network 1 and 2. The learning curves for network 1 and 2 are provided in the Supplementary.

\subsection{Post-processing}
The 30-dimensional VDL was reduced to 3 dimensions for visualization. This dimensional reduction was done using two methods: linear discriminant analysis (LDA) and principal component analysis (PCA). LDA is a supervised approach where the labels of the patients diagnosed using CC were used. LDA minimizes the distance between VDL points with the same label and maximizes the distance between points with different labels. Whereas PCA is an unsupervised approach that reduces the dimension by projecting the 30-dimensional VDL vectors onto 3 vectors that lie along the highest variances in data. We used an open source Python library Scikit-learn \cite{a52} to perform dimension reduction using LDA and PCA.
Additionally, we generated probability-based boundaries of the VDL points corresponding to different disease groups to quantify the distribution of various diseases as captured in the VDL. This is essentially a supervised classification task where a probability is assigned to every VDL point for each of the 12 disease groups shown in Table \ref{cohort_details}. This was done using a Random Forest classifier with the label for each 30-dimensional VDL vector taken from CC. For this training, only the original dataset was used and not the augmented ones since their labels were unknown. 25\% of the original dataset of 6244 were used for testing and the rest for training. We used Scikit-learn for the Random Forest analysis with 1000 estimators and achieved a Subset accuracy (or Exact match) of 0.74 on the test dataset. We also trained a Random Forest classifier to predict the probability of a point in the VDL to be peristaltic or not. The parameters of the classifier as well as the train-test split were the same as described above. Each data point was manually labelled as 1 for peristalsis or 0 for no peristalsis. The Jaccard score on this test set was 0.94. Further details on the accuracy are provided in the Supplementary.  The trained Random Forest classifiers add another capability to the MI-VAE framework to automate the diagnosis process in which it can predict an esophageal disorder as well its peristaltic behavior without manual intervention.

\section{Results and discussion}

\subsection{Virtual disease landscape in reduced dimensions}
\begin{figure}
  \centerline{\includegraphics[width=\textwidth]{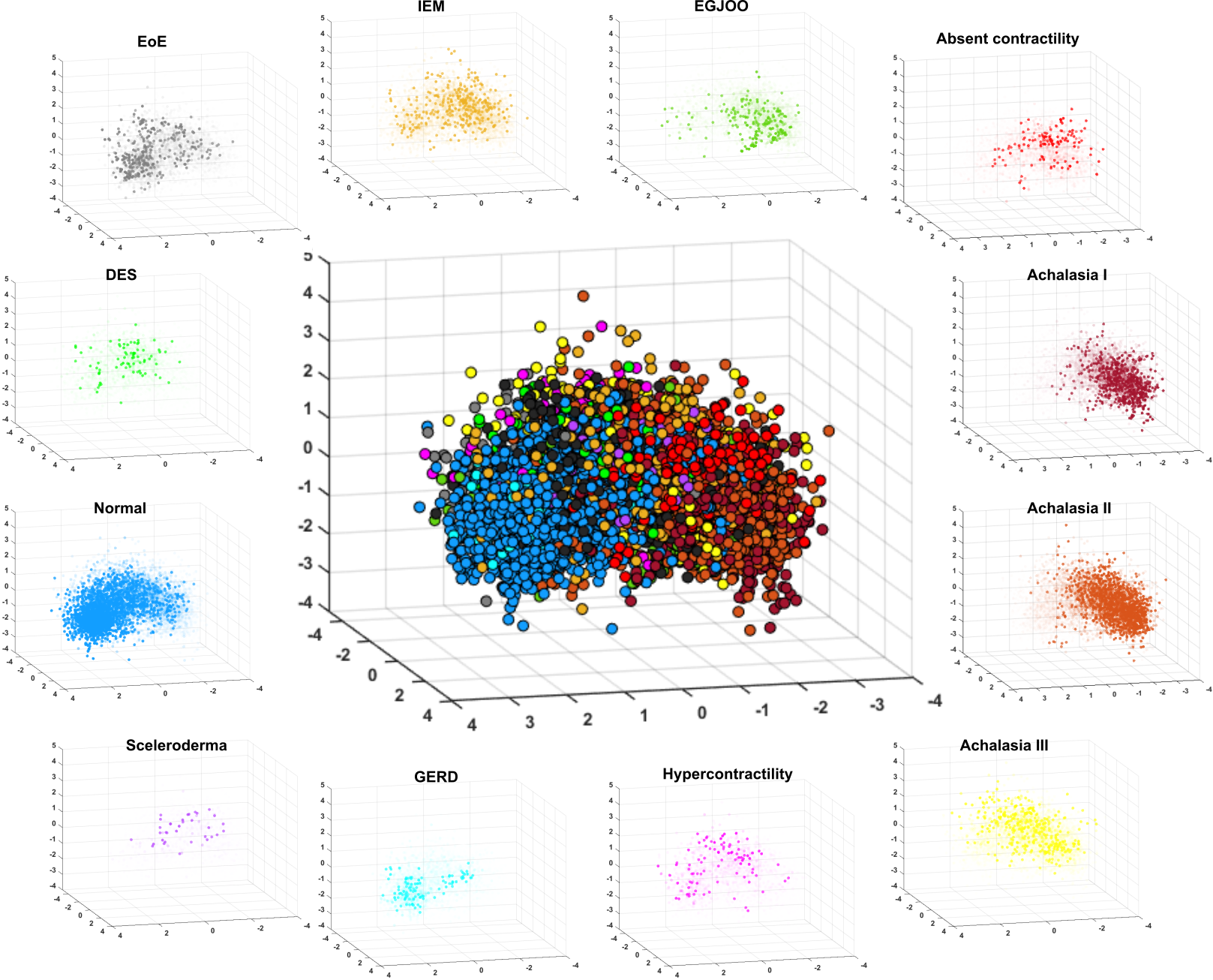}}
  \caption{Dimension reduced VDL using LDA. The different disease groups are shown with different colors and their probability is described by the intensity of the markers in the smaller plots.}
 \label{fig-lda}
\end{figure}
\begin{figure}
  \centerline{\includegraphics[width=\textwidth]{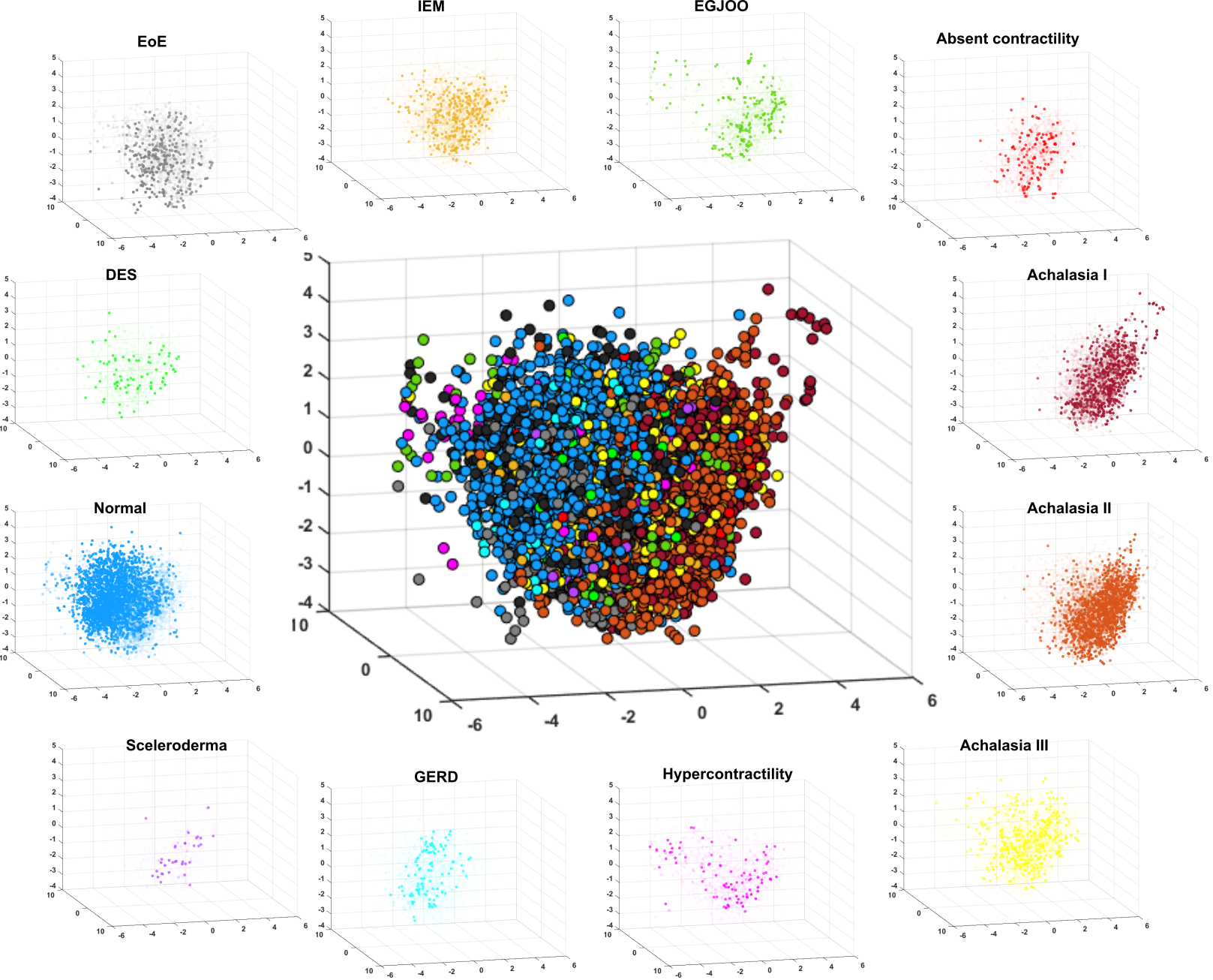}}
  \caption{Dimension reduced VDL using PCA. The different disease groups are shown with different colors and their probability is described by the intensity of the markers in the smaller plots.}
 \label{fig-pca}
\end{figure}

The 30-dimensional VDL consisted of the 24-dimensional latent variable generated by network 1, and the 6-dimensional vector constructed with the mechanics-based discrete parameters. The VDL was populated by points corresponding to the original dataset and not the augmented data since their labels are unknown. Figs. \ref{fig-lda} and \ref{fig-pca} show the dimension reduced VDL generated by LDA and PCA, respectively. The different diseases clustered into different regions of the VDL, but at the same time, parts of these clusters overlaped onto each other. As shown in Fig. \ref{fig-lda} with the dimension reduced VDL using LDA (ldaVDL), most of the normal subjects, GERD, and EoE patients lay on the left side of the VDL, whereas Achalasia, EGJOO, absent contractility patients lay on the right side of the VDL. The rest of the diseases were distributed midway between the two extremes. The overlap and separation between the different groups depend on their characteristic as observed on an EndoFLIP test. For instance, normal subjects and GERD often show similar behavior on the EndoFLIP with both exhibiting peristalsis as well as EGJ opening. On the other hand, Achalasia, EGJOO, and absent contractility patients show similar characteristics on the EndoFLIP with no contractile behavior and a closed EGJ. The other esophageal disorders that lay in between the two extremes of the VDL show a miscellaneous contractile behavior from weak peristalsis to irregular contractile response along with little to no EGJ opening. 
\begin{figure}
  \centerline{\includegraphics[width=\textwidth]{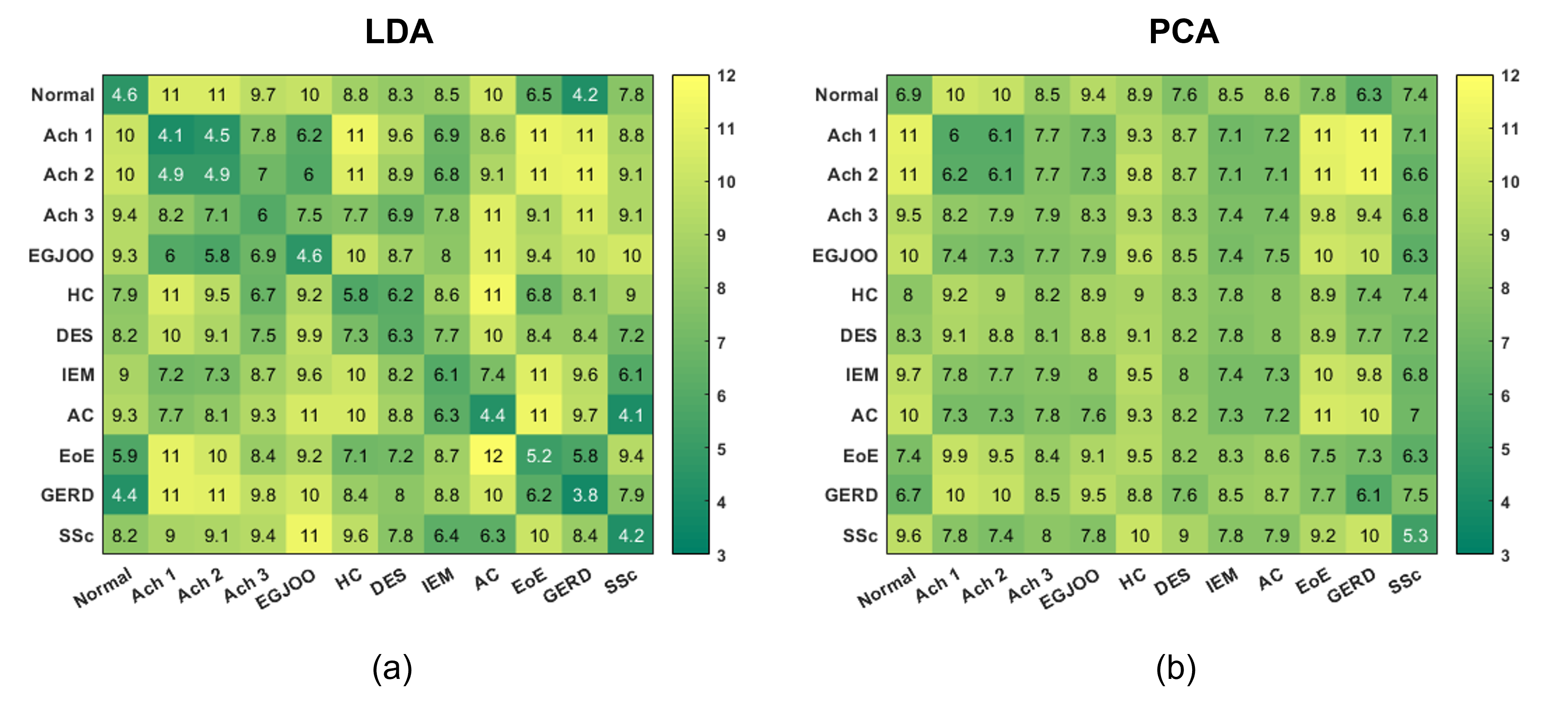}}
  \caption{Distance matrix showing the median distance between the points of each disease group specified by columns with the centroid of disease cluster specified by the rows. Each row has been normalized to represent distances as percentage and so each row adds up to 100.}
 \label{fig-distance_matrix}
\end{figure}
A similar behavior was also observed in the dimension reduced VDL using PCA (pcaVDL) as shown in Fig. \ref{fig-pca}. But the overlap between the different disorders in pcaVDL was more compared to that observed in ldaVDL. This can be seen more quantitatively in the distance matrix heatmaps of Fig. \ref{fig-distance_matrix}. An element in the i-th row and j-th column of the distance matrix is the median distance between the points of the j-th cluster and the centroid of the i-th cluster. Thus, for a good separation between the clusters, the diagonal elements should have smaller values compared to the off-diagonal elements. As shown in Fig. \ref{fig-distance_matrix}, the diagonal elements of distance matrix for ldaVDL were smaller than that of pcaVDL. Also, the off-diagonal elements of the distance matrix for ldaVDL were higher compared to that of pcaVDL. Therefore, as expected, the LDA performed better at segregating the different clusters since it is a supervised approach of separating the points based on their labels. But it should be noted that the 30-dimensional VDL was generated in an unsupervised manner. This high dimensional VDL captures the similar and dissimilar features of the input corresponding to the various disorders as well as other features necessary to generate a meaningful output. Thus, dimension reduction based on the variance of the data in an unsupervised manner as done in PCA provides insights about the structure of the 30-dimensional VDL. Therefore, the different clusters still observable after PCA showed that the VDL was successful is identifying the features that distinguish different esophageal disorders. The distance matrix as shown in Fig. \ref{fig-distance_matrix}(a) also quantifies the similarities that some disease groups share as can be seen qualitatively in Fig. \ref{fig-lda}. For instance, in row 1 which corresponds to normal subjects, normal and GERD had similar values. Similarly, in row 11 which corresponds to GERD, the median values for normal and GERD subjects had lower values. Again, in rows 2-3 which correspond to Achalasia Type I and Type II, the median distance values for these two disease groups were very similar and much lower than the other disease groups. Similar characteristics can be observed in Fig. \ref{fig-distance_matrix}(b), but not as prominent. Thus, the VDL captured both the similarities and dissimilarities between various esophageal disorders. The choice of dimension reduction depends on the application of the VDL. The pcaVDL should be chosen if there is low confidence in the labels of the data.

\subsection{Peristalsis represented in the VDL}
\begin{figure}
  \centerline{\includegraphics[width=\textwidth]{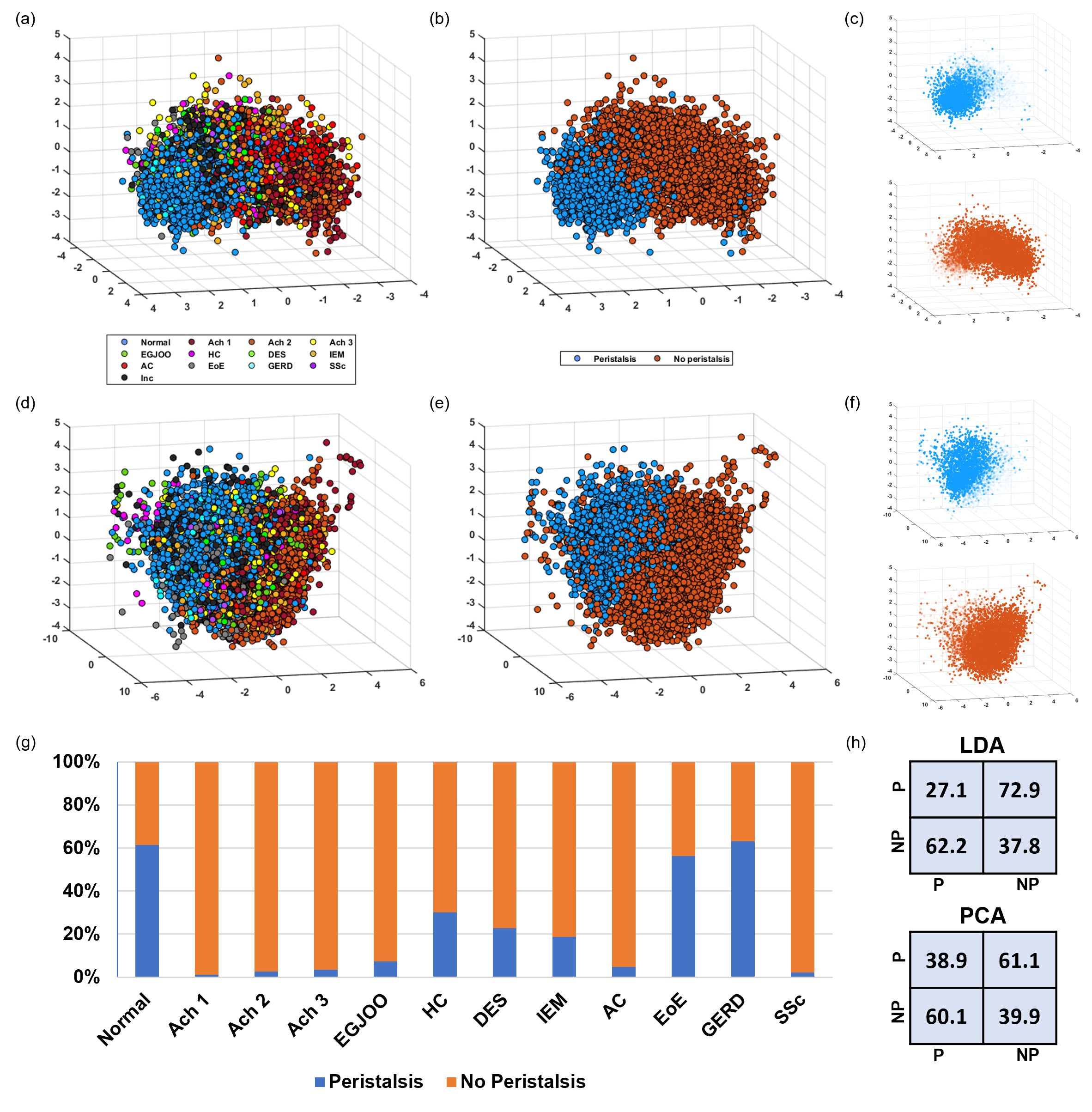}}
  \caption{Relating peristaltic behavior with the different disease groups. (a) LDA representation of the VDL with different disese groups (ldaVDL); (b) peristalsis and no peristalsis shown in ldaVDL; (c) Probability of every point from the dataset being whether peristaltic or not classified using Random Fores shown in ldaVDL; (d) PCA representation of the VDL with different disese groups (pcaVDL); (e) peristalsis and no peristalsis shown in pcaVDL; (f) Probability of every point from the dataset being whether peristaltic or not classified using Random Forest shown in pcaVDL; (g) Bar plots showing the percentage of $\theta$ variations of each disease group being either peristaltic or not peristaltic; (h) Distance matrices measuring the effectiveness of clustering for peristaltic behavior in ldaVDL and pcaVDL.}
 \label{fig-peristalsis}
\end{figure}
The labels or categories of different esophageal disorders as shown by the different colors in Figs. \ref{fig-lda} and \ref{fig-pca} correspond to the same subjects who had undergone manometry study. They were classified into the different disease categories based on CC. The classification labels were taken from the manometry study and not directly from EndoFLIP examination for two reasons. Firstly, a formal classification standard for EndoFLIP has not been developed yet while a formal classification standard exists for manometry through the CC. Secondly, some distinct esophageal disorders identified using manometry is not differentiable in EndoFLIP. For instant, normal and GRED subjects often show similar response on EndoFLIP, as do Achalasia Type I and Type II. The VDL developed using MI-VAE serves two important purposes regarding the points mentions above. Since the actual 30-dimensional VDL was developed in an unsupervised manner, the labels were not used directly. Therefore, the high dimensional clusters (which cannot be visualized directly) provide a quantitative estimate in classifying different esophageal disorders for EndoFLIP. Additionally, when the labels were used to visualize the VDL in reduced dimensions through ldaVDL, the VDL was able to quantitatively estimate the similarities and dissimilarities between the disease groups, and thus, provided an estimate of the capability as well as limitations of the EndoFLIP device in identifying the different categories of esophageal disorders.
Even though the colors used to represent the different esophageal disorders in a patient-specific manner provide valuable insights regarding the distribution of the different esophageal disorders on the VDL, the labels do not directly represent any specific characteristic of esophageal motility or esophageal wall properties. For instance, a normal subject might have proper peristalsis 70\% of the time and improper or no peristalsis for the rest. In such a case, all the points corresponding to this subject would be labelled the same irrespective of improper peristalsis in 30\% of them. Hence, the HRM label is not a fundamental physio-marker, while on the other hand, peristalsis is a fundamental physio-marker to distinguish esophageal disorders. Additionally, it is also possible that some subjects might respond differently on EndoFLIP when compared to HRM cite{a9}. We, therefore, investigated the peristaltic behavior of all the points that populate the VDL. Fig 10 shows the point cloud of ldaVDL and pcaVDL with labels indicating peristaltic behavior. A Random Forest classifier predicted the probability of a point in the VDL to be peristaltic or not. The probabilities of the points in the VDL are shown in reduced dimensions in Figs. \ref{fig-peristalsis}(c) and \ref{fig-peristalsis}(f) for ldaVDL and pcaVDL, respectively. The VDL points for peristalsis and without peristalsis were clustered very nicely without significant overlap, which is also evident from the high accuracy of the Random Forest classifier. This happened because the distinction between peristalsis and no peristalsis is a more fundamental criterion than the label imposed through manometry. The peristaltic behavior as identified by network 1 is an important hidden feature of $\theta$ variations, and therefore, the latent space variable captured this peristaltic behavior and led to the clustering as shown in Fig. \ref{fig-peristalsis}(b)-(c) and 10(e)-(f). The effectiveness of this clustering based on peristaltic behavior is presented more quantitatively in the distance matrices in Fig. \ref{fig-peristalsis}(h). Clearly there was a significant difference between the diagonal and off-diagonal values of the distance matrices, which signifies that the points in a particular group (peristalsis or no peristalsis) remain close to each other and far from the other group. The bar plot in Fig. \ref{fig-peristalsis}(g) quantitatively estimates what percentage of each disease group exhibit peristalsis. As already described above, most of the normal, GERD and EoE subjects have peristaltic behavior with close to 60\% of the $\theta$ variations show peristalsis, while the Achalasia patients have more than 90\% of the $\theta$ variations without any peristalsis. 

\subsection{Continuous behavior of the VDL}
\begin{figure}
  \centerline{\includegraphics[width=\textwidth]{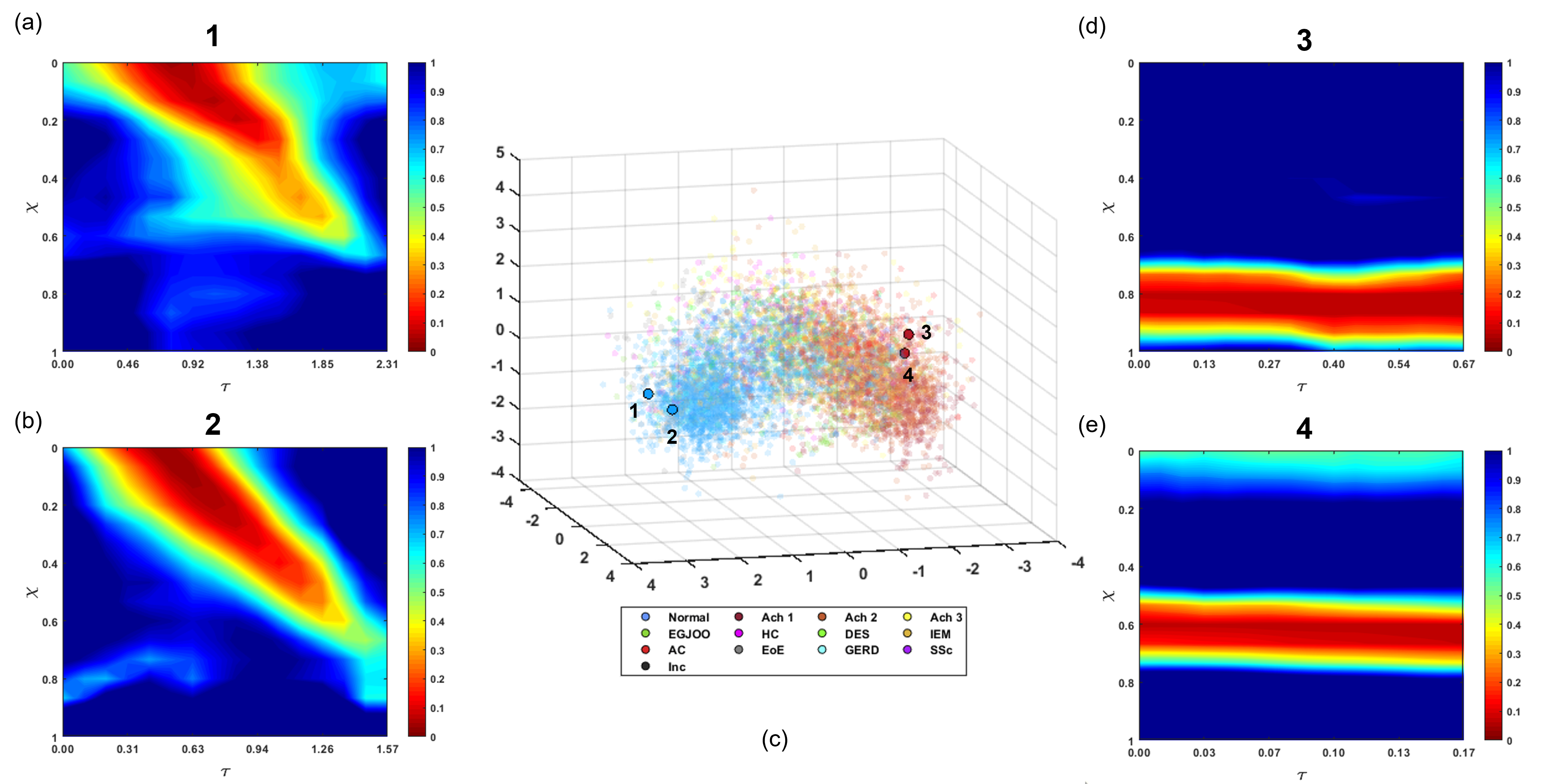}}
  \caption{The continuous behavior of the VDL is shown through points selected from its two extremes: Normal subjects and Achalasia Type I and II patients. (a) and (b) show a typical contraction pattern in normal subjects with strong peristalsis and relaxed EGJ. (d) and (e) show the characteristics of typical Achalasia patients with no contraction at the esophageal body and a strong tone at the EGJ marked by the horizontal red zone.  (c) VDL represents of the contraction patterns 1-4.}
 \label{fig-vae_extremes}
\end{figure}
The capability of the VDL to identify the similarities and dissimilarities between esophageal disorders is further observed in Fig. \ref{fig-vae_extremes}. Two close points were selected from the extreme left end of the normal subjects and two close points were selected from the extreme right end of the Achalasia Type II. The two points selected from the controls had similar characteristics since they lay close to each other. Both showed strong peristaltic behavior (seen by the oblique red band) and a relaxed EGJ (no tone at the distal esophagus). Similarly, the two points of Achalasia Type II showed similar variation of the activation parameter. Both these points displayed no contraction and the EGJ remained closed as shown by the horizontal red band at the distal esophagus. Since the points 1 and 2 were far from the points 3 and 4, they displayed completely different behavior. On comparing the variation of the activation parameter for points 3 and 4, we can see that the location of the EGJ were different for the two cases. The EGJ was much higher in case 4 compared to case 3. This difference occurs because the EndoFLIP is manually positioned at different location during examination. But the points 3 and 4 lay very close to each other on the VDL. Thus, we conclude that the VDL visualized in reduced dimensions using LDA is not sensitive to the placement of the EndoFLIP. This characteristic removes any error in our prediction due to differences in the manual use of the EndoFLIP device. Therefore, the VDL describes a parameter space dependent only on the state and functioning of the esophagus and not on way in which measurements are taken using the EndoFLIP.

\subsection{Generative property of the MI-VAE}
\begin{figure}
  \centerline{\includegraphics[width=\textwidth]{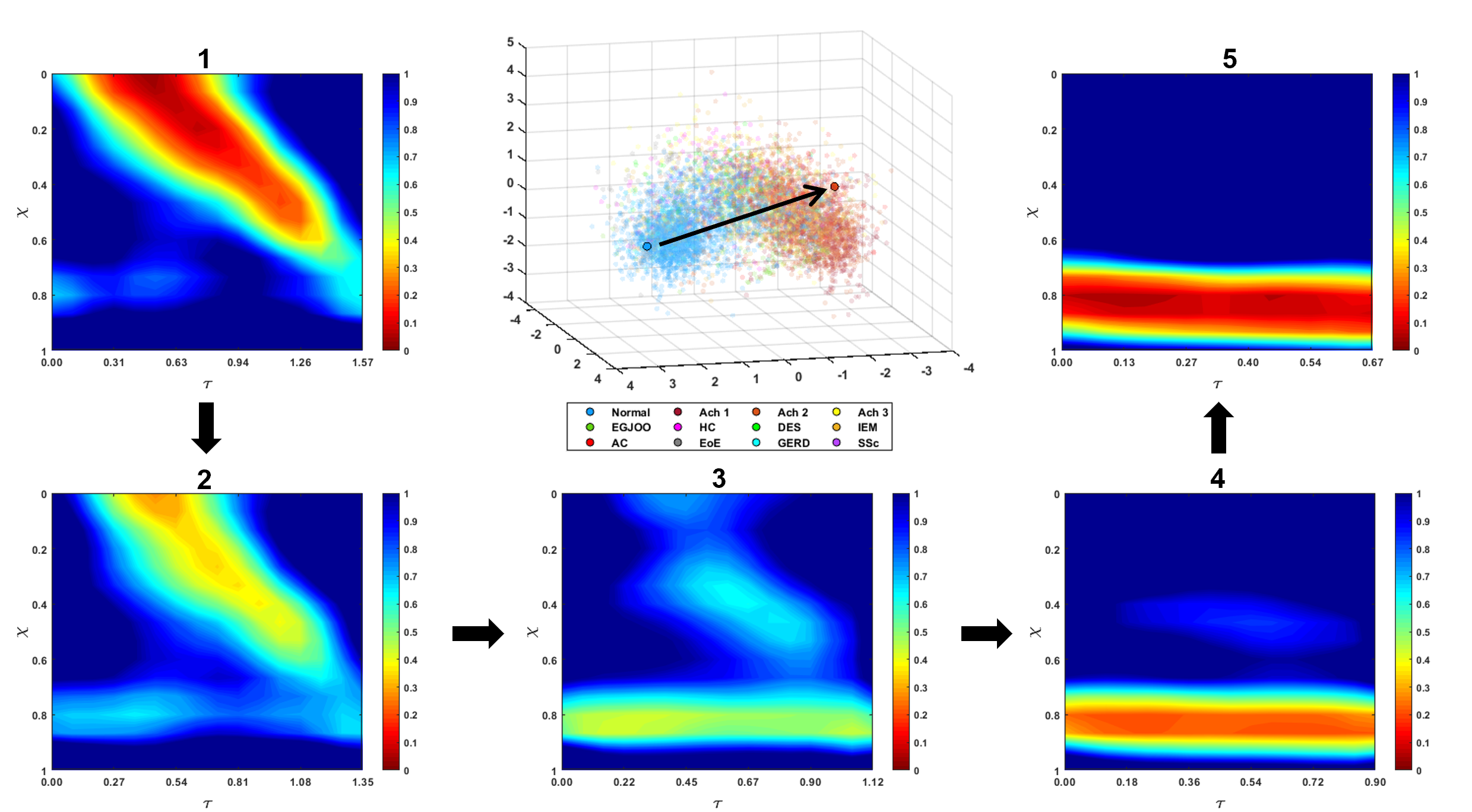}}
  \caption{Example to describe the continuous nature of the VDL. Two points 1 and 5 are chosen at the extremes of normal subjects and Achalasia Type II patients, respectively. Points 2-4 are chosen in sequence to divide the vector joining 1 and 5 in four equal parts.  The $\theta$ variations generated for 2-4 show the transition from normal to Achalasia characteristics.}
 \label{fig-z_traverse}
\end{figure}
One of the most important features of the MI-VAE is its generative capability. Due to the continuous nature of the VDL, new vectors from the VDL i.e., those which were not present in the training set can generate meaningful representations of the mechanics-based parameters. Fig. \ref{fig-z_traverse} shows how transition from one point at one extreme of the normal cohort to another point in the other extreme of Achalasia Type II can lead to different generations. A 30-dimensional vector was calculated between these two points and 3 points were chosen along this vector dividing the vector in 4 equal parts. Thus, the 2 end points were known but the 3 points in between were new and generated mechanics-based parameters not known before. As already described before, the first point taken from the normal cohort exhibited peristalsis and a relaxed EGJ, and the last point in the Achalasia Type II zone exhibited no peristalsis and a strong unrelaxed EGJ tone. As the transition occurs between these 2 points, we see from the 3 generated outputs that the peristalsis strength gradually weakens and the EGJ tone gradually strengthens. 

Additionally, we can also retrieve all the other mechanics-based parameters apart from the $\theta$ variation which are $K/A_o$, $P_{\mathrm{max}}$, $T_{\mathrm{max}}$, $\theta_{\mathrm{max}}$, and the three EGJ work metrics EGJW, $EGJROW_1$ , $EGJROW_2$, and $EGJROW_3$ through network 2. This is the main reason for choosing mechanics-based parameters as the input of the MI-VAE instead of the raw distal pressure and cross-sectional area variations. The generated parameters have physical significance and directly estimates the mechanical health of the organ unlike the raw data generated from EndoFLIP which do not directly predict the state and functioning of the esophagus. Additionally, with the use of the mechanics, the two distributed measurements: cross-sectional area, $A(x,t)$, and distal pressure $P_d (t)$, were combined into one activation parameter $\theta(x,t)$, which not only makes it easy to apply the MI-VAE, but also has a physical meaning since it estimates the shortening of the esophageal muscle fiber lengths.
With this framework, it is possible to track the progression of an esophageal disorder with time and using this time-series data, it is possible to extrapolate in the VDL and predict the future state of the esophagus. For instance, if the contraction pattern of a subject is found to move from 1 to 2 (shown in Fig. \ref{fig-z_traverse}) in the first year, and from 2 to 3 in the second year, then it can be extrapolated that the state of the esophagus most likely would be as shown by the contraction pattern 4 in the third year. Thus, the MI-VAE not only describes the current state of the esophagus, but also predicts the disease progression. 

\subsection{Application in estimating effectiveness of a treatment}
One of the major usages of the EndoFLIP device is to accurately diagnose Achalasia patients. Inflating the EndoFLIP bag at the EGJ tests the resistance against esophageal emptying due to EGJ tone as well as the capability of the contraction to force the EGJ to open. The treatment of an Achalasia patient essentially involves reducing the EGJ tone. Usually, this is done through the usage of topical steroid for less severe cases \cite{a53,a54,a55} and myotomy procedures such as Laparoscopic Heller myotomy (LHM) \cite{a56,a57} and Peroral Endoscopic Myotomy (POEM) \cite{a58,a59} for severe cases. Usually, FLIP is performed before and after these treatment procedures to test the effectiveness of a treatment as well as for tracking the esophageal condition for multiple years after treatment. In the next two subsections, we discuss two such scenarios where the MI-VAE framework can be applied to aid EndoFLIP diagnosis.

\subsubsection{Pre- and post-treatment state of the esophagus in Achalasia patients}
\begin{figure}
  \centerline{\includegraphics[width=\textwidth]{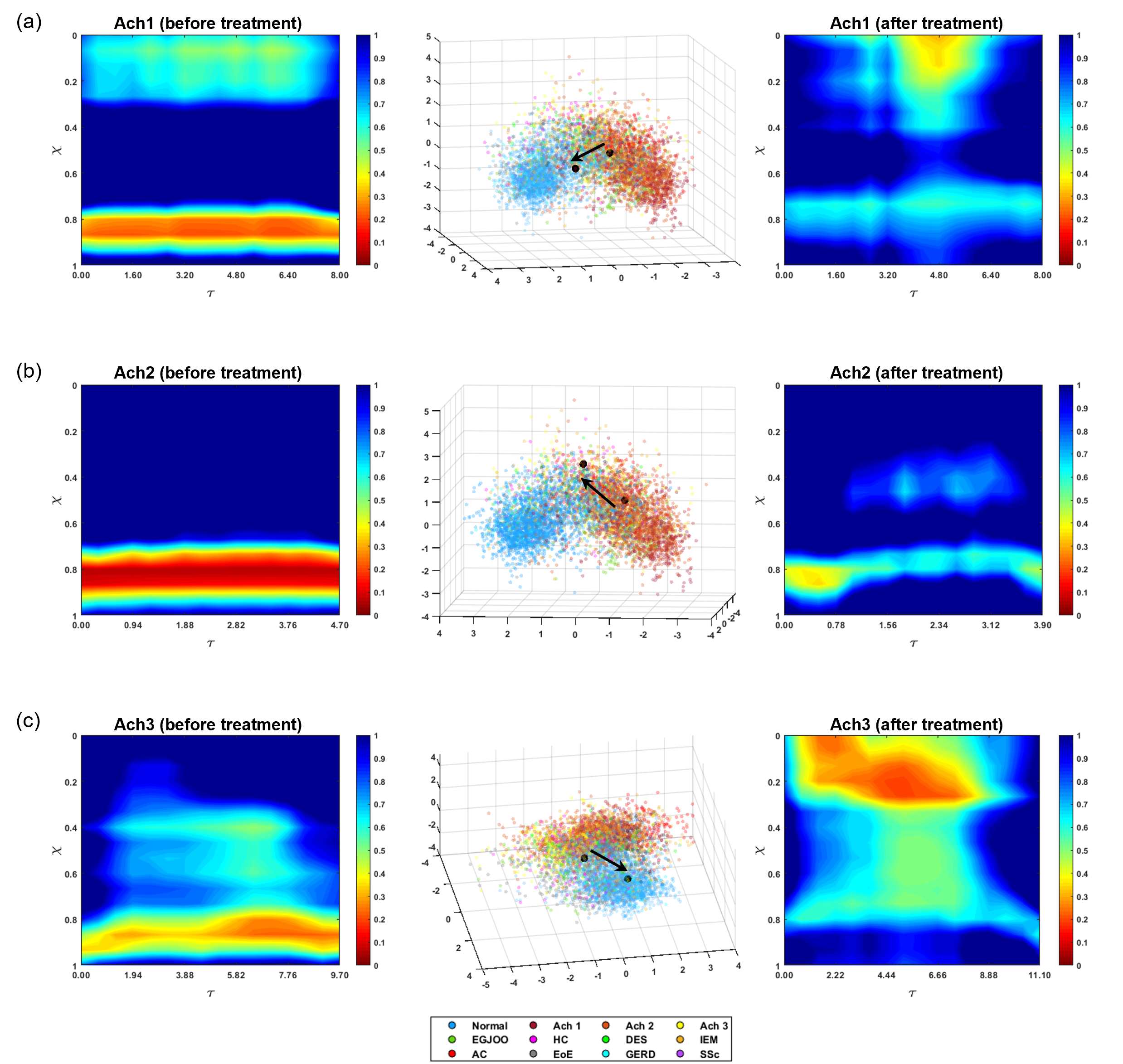}}
  \caption{Estimating the effectiveness of treatment with steriods on Achalasia patients using MI-VAE. (a)-(c) show the $\theta$ variation before and after treatment with steroids in Achalasia Type I, II, and III, respectively. The contour plots on the left show the behavior before treatment and the figures on the right show the behavior after treatment.  The figures in the center show the VDL representations of the contour plots. The arrow shows the direction from before to after the treatment.}
 \label{fig-ach_before_after}
\end{figure}
Steroid therapy is a popular treatment for Achalasia as a first step and often it is quite effective for minor cases. A quantitative assessment of the effectiveness of a topical steroid treatment is therefore very important. Using the MI-VAE framework, we present a quantitative approach for such an assessment. We tested this on three Achalasia patients, one for each Achalasia Type. Fig. \ref{fig-ach_before_after} shows the contraction patterns before and after steroid treatment as well as their placement on the VDL. We selected the “best” $\theta$ variation for each case by identifying the most effective peristalsis and the least EGJ tone. All these contraction patterns were observed at the same EndoFLIP bag volume of 50 mL. In Fig. \ref{fig-ach_before_after}(a), we see that for the Achalasia Type I patient, after steroid treatment, the EGJ tone had decreased significantly with minute improvement in contraction strength. On the VDL, this corresponded to the movement from Achalasia zone towards the normal subjects. A similar observation was also made for the Achalasia Type II patient as shown in Fig. \ref{fig-ach_before_after}(b). Since the peristalsis had not been recovered and the EGJ did not relax fully, the final point after topical steroid treatment did not lie completely in the normal cohort zone. But the shift towards the normal cohort does indicate improvement.  In Fig. \ref{fig-ach_before_after}(c), we see that for the Achalasia Type III patient, with steroid treatment the contraction strength had improved significantly (although not fully peristaltic) and the EGJ had relaxed as well. This behavior is very similar to that observed in normal subjects, and so, we see that the corresponding point in the VDL lay more in the normal cohort zone than that seen in Figs. \ref{fig-ach_before_after}(a) and \ref{fig-ach_before_after}(b). It is not possible to identify whether the EGJ relaxed on its own or due to the pressure developed by the contraction at the esophageal walls. But it matches observations as typically seen in normal subjects. The improvement after treatment can be estimated quantitatively by the magnitude of the vector drawn from the initial to the final point in the VDL. The direction of the vector quantitatively estimates the direction of improvement. This vector corresponds not only to the $\theta$ variation, but also the discrete mechanics-based parameters. This adds physical meaning to the quantitative assessment to the effectiveness of the treatment. 

\subsubsection{Post-POEM tracking of esophageal condition}
\begin{figure}[ht]
  \centerline{\includegraphics[width=\textwidth]{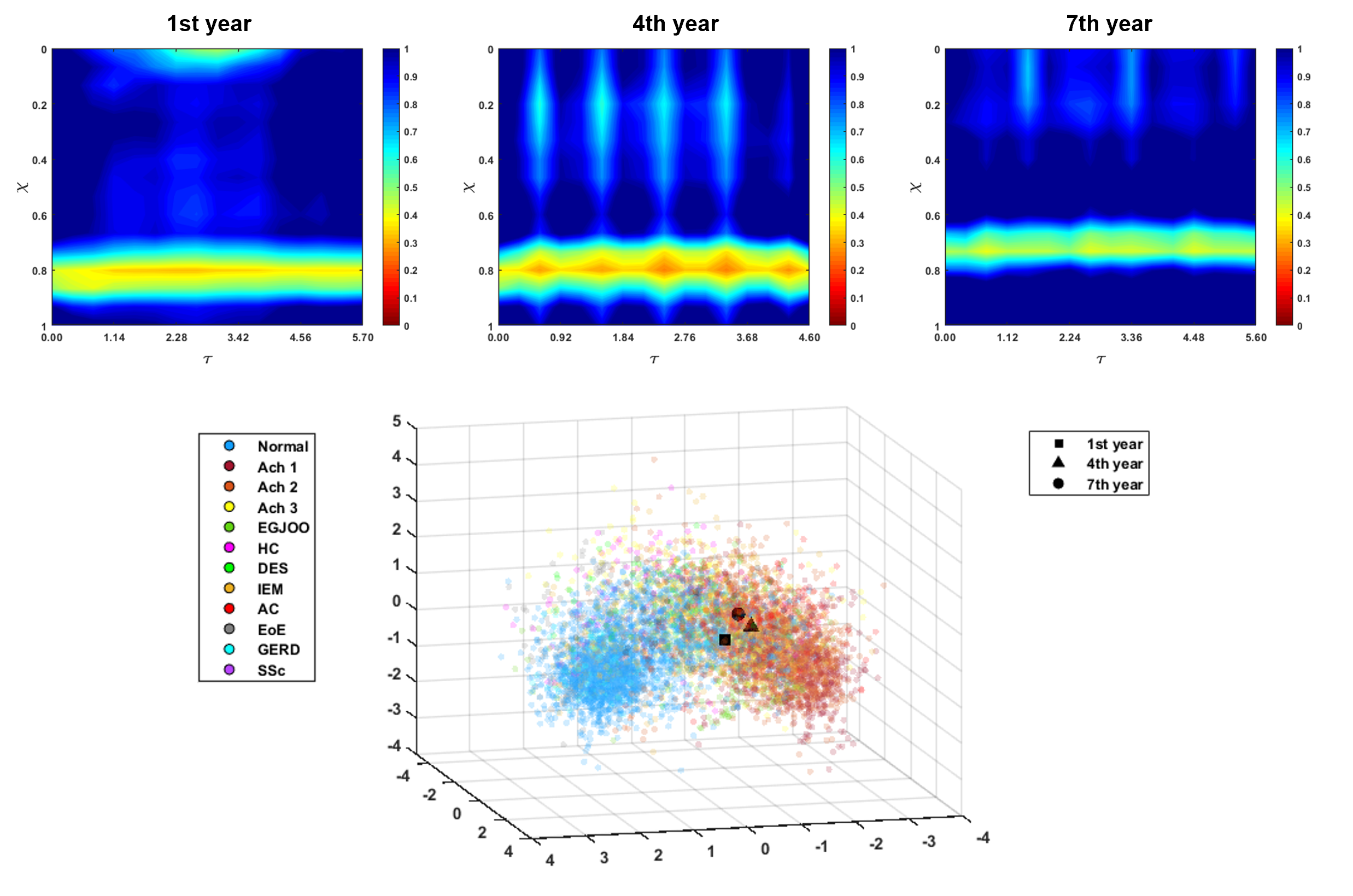}}
  \caption{Tracking the state and motility of the esophagus after POEM for 7 years. Three $\theta$ variations are shown as observed in the 1st, 4th, and 7th year. These are plotted in their VDL representation. The three points lie close to each other indicting that the patient has stable esophageal motility charcateristics for the 7 years they have been tracked.}
 \label{fig-post_poem}
\end{figure}
Myotomy is a surgical procedure performed as a last resort to treat achalasia patients. In this procedure, the esophageal muscle fibers are broken at the lower esophageal sphincter (LES) to weaken the inherent tone at the EGJ. This makes it easier for swallowed food to empty from the esophagus. POEM is a type of myotomy procedure where the circular muscle fibers are broken endoscopically at the LES. Once this procedure has been completely, it is often important to keep track of the patient condition to check if the EGJ tone has recovered with time leading to achalasia symptoms or other any other complications like blown-out myotomy (BOM) \cite{a60} have developed. We tested our MI-VAE framework by tracking the condition of one patient who had undergone POEM. The patient FLIP data was available for the first year, fourth year, and seventh year after POEM. The $\theta$ variations for each of these three years are shown in Fig. \ref{fig-post_poem}. The three plots look very similar with a weakened EGJ tone compared to the typical Achalasia cases discussed above. Although, peristalsis had not recovered in those years, EGJ tone did not deteriorate as well. Therefore, we see the corresponding VDL plot for those $\theta$ variations to lie close to each other and indicated that the patient condition was stable over the years. Since there was no observable peristalsis but the weak EGJ tone remained, we see the three points lay midway between the two extremes of normal and Achalasia cohorts. This example shows how the MI-VAE framework can be used for post-POEM tracking of patient condition but can be applied very easily to other treatment procedures as well.

\subsection{Limitations}
The MI-VAE framework provides an effective technique to map different esophageal disorders onto a parameter space (called the VDL) based on their mechanical characteristics as well as predict the mechanical ‘health’ of the esophagus, aid in disease diagnosis and treatment.  But it also has some limitations. Firstly, the raw data output from the EndoFLIP cannot be used directly by this framework. The MI-VAE requires the manual identification of the time instants between which readings are to be considered. This manual intervention might introduce differences in prediction based on which time instants are chosen. Secondly, the EndoFLIP device has some technical limitations. It has an upper measurement limit for diameter which corresponds to the maximum distension of the EndoFLIP bag. Strong contractions might sometimes lead to the bag distending higher that this upper limit and cause incorrect readings.  It also a lower limit for its measurements which is close to the catheter diameter. Again, strong contractions might cause full collapse of the EndoFLIP bag on the catheter leading to incorrect readings. Additionally, it is sometimes observed that the bag volume calculated using the diameter readings might not be equal to the actual recoded bag volume. These factors might cause errors in the prediction of MI-VAE. Thirdly, all the esophageal disorders are not well represented in the dataset. The characteristics of the disorders represented by a smaller dataset (like Scleroderma) might not be learned properly by the MI-VAE. Therefore, their prediction, especially with the Random Forest classifier, might not be as accurate as the disorders represented by a larger dataset. Fourthly, reduction of the VDL dimensions using LDA and PCA for visualization might lead to loss to important features that define the state and functioning of the esophagus. Lastly, as described earlier, the labels used for dimension reduction using LDA as well as training the random forest classifier is patient-specific, and not specific to the mechanics-based parameters. For instance, some $\theta$ variations of normal subjects might not exhibit peristaltic behavior. This might introduce some errors in the prediction of MI-VAE.    

\section{Conclusion}
In this work, we presented a framework called mechanics-informed variational autoencoder (MI-VAE) that quantitatively identified and distinguished the various esophageal disorders based on their physical characteristics through a parameter space called the virtual disease landscape (VDL). The physical characteristics were estimated through a set of physical parameters such as esophageal wall stiffness, contraction pattern, active relaxation of the esophageal wall muscles, and work metrics estimating EGJ behavior. These parameters were solved in a patient-specific manner from EndoFLIP data using a one-dimensional mechanics-based inverse model. The VDL identified the similarities and dissimilarities between the various esophageal disorders as well as classified them based on their peristaltic behavior. Additionally, Random Forest classifiers trained on the data represented in the VDL add a predictive capability to this framework to automatically identify different esophageal disorders as well as their peristaltic behavior. We also described how the generative property of the MI-VAE gives it the capability to predict disease progression in time. Finally, we demonstrated through clinical applications that the MI-VAE can estimate the effectiveness of a treatment as well as track patient condition after treatment. Although this work focusses on the use of data from EndoFLIP, it can very easily be extended to other diagnostic technologies as well as other organs. With all the capabilities of the MI-VAE, it provides an effective framework to not only gain fundamental insights about the functions of an organ, but also prove valuable in disease diagnosis and treatment.  

\section*{Acknowledgments}

This research was supported in part through the computational resources and
staff contributions provided for the Quest high performance computing facility
at Northwestern University which is jointly supported by the Office of the
Provost, the Office for Research, and Northwestern University Information
Technology.

This work also used the Extreme Science and Engineering Discovery Environment
(XSEDE) clusters Comet, at the San Diego Supercomputer Center (SDSC) and
Bridges, at the Pittsburgh Supercomputing Center (PSC) through allocation
TG-ASC170023, which is supported by National Science Foundation grant number
ACI-1548562 \cite{xsede}.

\section*{Funding Data}
\begin{itemize}
\item National Institutes of Health (NIDDK grants DK079902 \& DK117824; Funder ID: 10.13039/100000062)
\item National Science Foundation (OAC grants 1450374 \& 1931372; Funder ID: 10.13039/100000105)
\end{itemize}

\bibliographystyle{elsarticle-num}
\bibliography{mi_vae}
\newpage

\section*{Supplementary}
\setcounter{figure}{0}
\makeatletter 
\renewcommand{\thefigure}{\@Roman\c@figure}
\makeatother
\subsection*{Numerical Implementation}
The non-dimensional mass and momentum conservation equations as described by Eqns. \ref{eqn-nd_continuity} and \ref{eqn-nd_momentum} are solved to calculate the fluid velocity and pressure, respectively. Eqns. \ref{eqn-nd_continuity} and \ref{eqn-nd_momentum} can be represented in terms of flow rate as shown below:
\begin{align}
 \frac{\partial \alpha}{\partial \tau} + \frac{\partial q}{\partial \chi} =0, \tag{S1} \label{eqn-nd_continuity1} \\
 \frac{\partial q}{\partial \tau} + \frac{\partial}{\partial \chi}\left(\frac{q^2}{\alpha}\right)+\alpha\frac{\partial p}{\partial \chi} + \varphi\frac{q}{\alpha} = 0, \tag{S2} \label{eqn-nd_momentum1}
\end{align} 
where, $q=u\alpha$ is the mean flow rate. Since the EndoFLIP is closed at its two ends, it is necessary to enforce zero flow rate boundary conditions at $\chi=0$ and $\chi=1$. Since Eq. \ref{eqn-nd_continuity1} requires only one boundary condition for $q$, we differentiate it with respect to $\chi$ to obtain a second order form as follows: 
\begin{align}
 \frac{\partial^2 \alpha}{\partial \tau \partial \chi} + \frac{\partial^2 q}{\partial \chi^2} =0. \tag{S3} \label{eqn-nd_continuity2}
\end{align} 
We do the same for Eq. \ref{eqn-nd_momentum1} to specify a Dirichlet boundary condition for pressure at the distal end (as measured by the pressure sensor) and zero pressure gradient at the proximal end as typically observed in practice. Eq. \ref{eqn-nd_momentum1} takes the following form after differentiating with respect to $\chi$:
\begin{align}
 \frac{\partial^2 q}{\partial \tau \partial \chi} + \frac{\partial^2}{\partial \chi^2}\left(\frac{q^2}{\alpha}\right)+\frac{\partial}{\partial \chi}\left(\alpha\frac{\partial p}{\partial \chi}\right) + \frac{\partial}{\partial \chi}\left(\varphi\frac{q}{\alpha}\right) = 0. \tag{S4} \label{eqn-nd_momentum2}
\end{align} 
\begin{figure}[ht]
  \centerline{\includegraphics[scale=1.2]{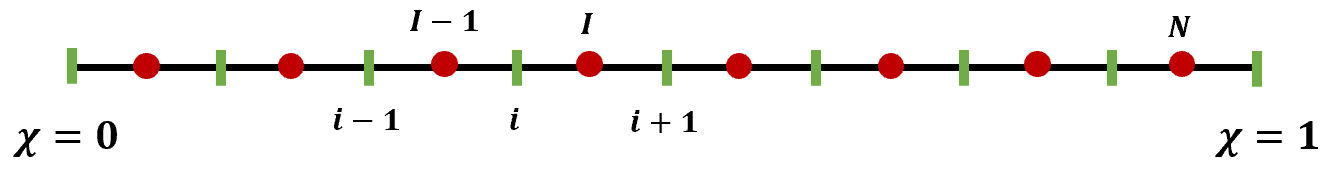}} 
  \caption{Schematic of the discretized domain. The green markers show the cell boundaries, and the red markers show the cell centers.}
  \label{fig-discretization}
\end{figure} 
Eqns. \ref{eqn-nd_continuity2} and \ref{eqn-nd_momentum2} are solved using the finite volume method on a grid as shown in Fig. \ref{fig-discretization}. 
The discretized form of Eqns. \ref{eqn-nd_continuity2} and \ref{eqn-nd_momentum2} are shown below:
\begin{align}
q_i - \frac{1}{2}\left(q_{i+1}+q_{i-1}\right) &= \frac{\Delta \chi}{2\Delta \tau}\left(\alpha_{I}-\alpha_{I-1}-\alpha_{I}^o+\alpha_{I-1}^o\right), \tag{S5} \\
\left(\alpha_{I}+\alpha_{I-1}\right)p_i - \alpha_{I}p_{i+1} - \alpha_{I-1}p_{i-1} &= \frac{\Delta \chi}{2\Delta \tau}\left(q_{i+1}-q_{i-1}-q_{i+1}^o+q_{i-1}^o\right) + \left(\frac{q_{i+1}^2}{\alpha_{i+1}}+\frac{q_{i-1}^2}{\alpha_{i-1}}-\frac{2q_i^2}{\alpha_i}\right)+\frac{\varphi}{2}\left(\frac{q_{i+1}}{\alpha_{i+1}}-\frac{q_{i-1}}{\alpha_{i-1}}\right), \tag{S6}
\end{align}
where, $i,I=1,2,…,N$. The non-dimensional cross-sectional areas $\alpha$ were known through the impedance sensors. 
\subsection*{Training details of MI-VAE}
\begin{figure}
  \centerline{\includegraphics[scale=0.5]{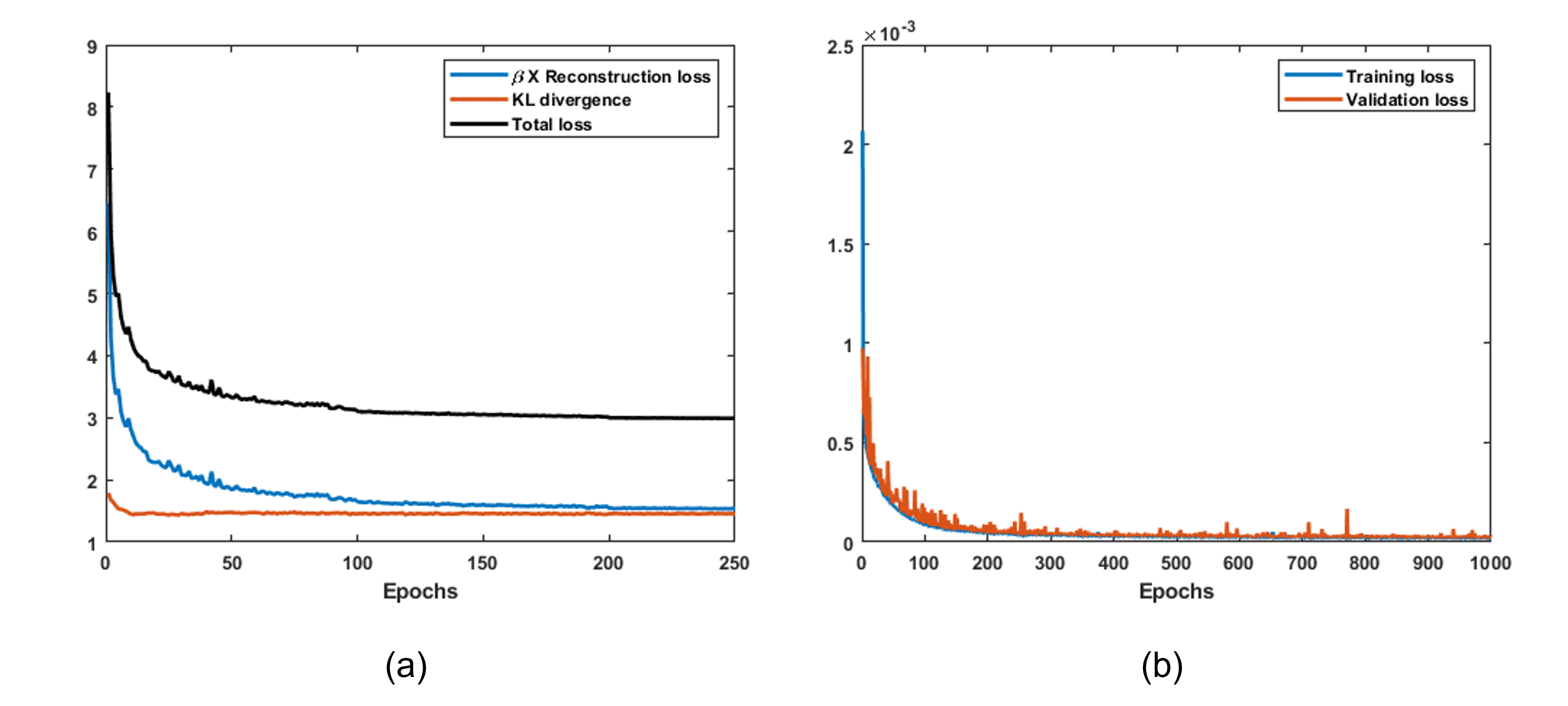}}
  \caption{Training details of MI-VAE. a) Learning curve for network 1, b) Learning curve for network 2.}
 \label{fig-learning_curves}
\end{figure}
The learning curves of network 1 and 2 are shown in Fig. \ref{fig-learning_curves} (a) and (b), respectively. Because of the scaling parameter $\beta$ as described Eq. \ref{eqn-loss_function}, the KL divergence and the product of $\beta$ and reconstruction loss have similar magnitude. If there was a significant difference between their magnitudes, the total loss would be represented mainly by the larger of the two and proper minimization of both the loss would not happen. The final magnitudes of reconstruction loss, KL divergence, and the total loss after 250 epochs were $1.36\times10^{-3}$, 1.54, and 2.9, respectively. The training and validation losses for network 2 (with learning rate $10^{-3}$) converge to $1.36\times10^{-5}$ and $1.88\times10^{-5}$, respectively.
\subsection*{Training details of the Random Forest classifiers}
\begin{figure}[ht]
  \centerline{\includegraphics[scale=0.5]{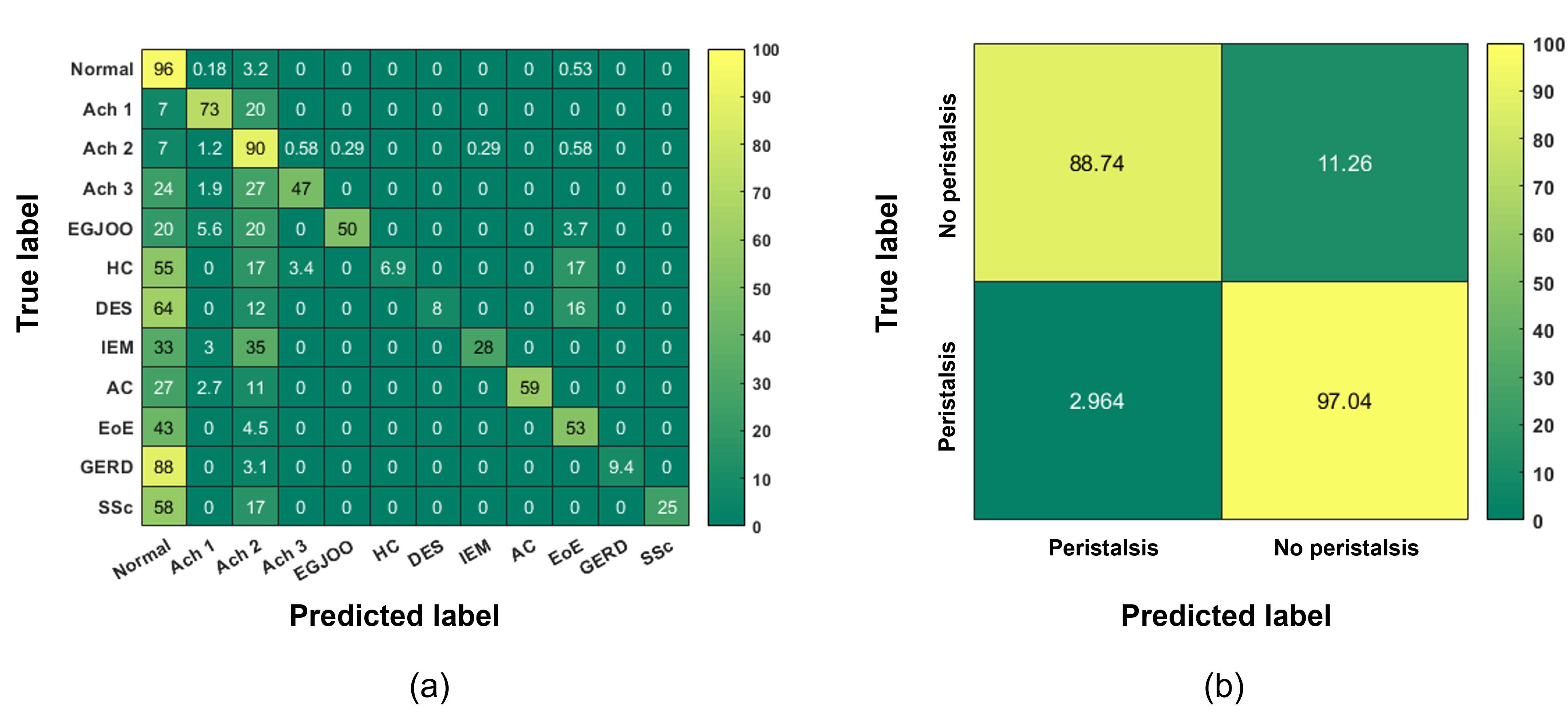}}
  \caption{Confusion matrices for Random Forest classifiers. a) Classifier for disease groups, b) Classifier for peristaltic behavior.}
 \label{fig-confusion_matrices}
\end{figure}

The confusion matrices corresponding to the two random forest classifiers are shown in Fig. \ref{fig-confusion_matrices}.  The numbers in the matrices are represented in terms of percentages. The classifier for the different disease groups (as shown in Fig. \ref{fig-confusion_matrices}(a)) shows reasonable accuracy for most groups. As already discussed, some groups exhibit very similar behavior on EndoFLIP although they were classified as different groups in through HRM. Additionally, a subject might have a variety of contraction patterns. This reduces the overall accuracy of the Random Forest classifier to some extent. The final Subset accuracy observed on the test set was 0.74.  The Random Forest classifier to identify peristaltic behavior performs better as shown by the confusion matrix in Fig. \ref{fig-confusion_matrices}(b). This is because the labels for peristaltic behavior were specified for each data point (corresponding to each $\theta$ variation) and labeled just by subject diagnosis. Also, there is less overlap in the VDL between the two groups, i.e., peristalsis and no peristalsis. The Jaccard accuracy observed on the test set was 0.94.
\end{document}